\documentclass[10pt,twocolumn,letterpaper]{article}

\usepackage{cvpr}              

\usepackage{booktabs}
\usepackage{multicol}
\usepackage{colortbl}
\usepackage{multirow}
\usepackage{amsthm}

\usepackage{amssymb}
\usepackage{pifont}
\usepackage{comment}
\usepackage[normalem]{ulem}
\usepackage{xcolor}
\usepackage{algorithmic}
\usepackage{algorithm}
\usepackage{wrapfig}
\usepackage{amsthm,amsmath,amssymb,,bm,bbm}
\usepackage{mdframed}
\usepackage{multicol} 
\theoremstyle{plain}
\theoremstyle{definition}
\usepackage{makecell}
\newmdtheoremenv{prop}{Proposition}

\newmdtheoremenv{definition}[theorem]{Definition}

\theoremstyle{remark}
\newmdtheoremenv{corollary}{Corollary}[theorem]
\definecolor{cvprblue}{rgb}{0.21,0.49,0.74}
\usepackage[pagebackref,breaklinks,colorlinks,allcolors=cvprblue]{hyperref}

\def\confName{CVPR}
\def\confYear{2026}

\title{LLaDA-MedV: Exploring Large Language Diffusion Models for Biomedical Image Understanding}

\author{Xuanzhao Dong$^{1*}$ \quad Wenhui Zhu$^{1*}$ \quad Xiwen Chen$^{2,5*}$ \quad Zhipeng Wang$^{3*}$ \quad Peijie Qiu$^{4}$ \quad Shao Tang$^{3}$  \\ \quad Xin Li$^{1}$ \quad Yalin Wang$^{1}$ \\
\\
$^{1}$ Arizona State University, AZ, USA, $^{2}$ Clemson University, SC, USA \\ 
$^{3}$ LinkedIn Corporation, CA, USA, $^{4}$ Washington University in St. Louis, MO, USA\\
$^{5}$ Morgan Stanley, NY, USA
}

\begin{document}
\maketitle
\def\thefootnote{*}\footnotetext{These authors contributed equally to this paper.}

\begin{abstract}
Autoregressive models (ARMs) have long dominated the landscape of biomedical vision-language models (VLMs). Recently, masked diffusion models such as LLaDA have emerged as promising alternatives, yet their application in the biomedical domain remains largely underexplored. To bridge this gap, we introduce \textbf{LLaDA-MedV}, the first large language diffusion model tailored for biomedical image understanding through vision instruction tuning. LLaDA-MedV achieves relative performance gains of 7.855\% over LLaVA-Med and 1.867\% over LLaDA-V in the open-ended biomedical visual conversation task, and sets new state-of-the-art accuracy on the closed-form subset of three VQA benchmarks: 84.93\% on VQA-RAD, 92.31\% on SLAKE, and 95.15\% on PathVQA. Furthermore, a detailed comparison with LLaVA-Med suggests that LLaDA-MedV is capable of generating reasonably longer responses by explicitly controlling response length, which can lead to more informative outputs. We also conduct an in-depth analysis of both the training and inference stages, highlighting the critical roles of initialization weight selection, fine-tuning strategies, and the interplay between sampling steps and response repetition. The code and model weight is released at \url{https://github.com/LLM-VLM-GSL/LLaDA-MedV}.

\end{abstract}

\section{Introduction}
Building on the success of general domain vision-language models (VLMs), researchers have increasingly shifted their focus to the biomedical domain, leading to the development of biomedical VLMs that have demonstrated notable progress across a variety of medical tasks~\cite{bannur2023learning,wang2022medclip,lin2023pmc}. Notably, autoregressive models (ARMs), known for their effectiveness in modeling multimodal distributions, currently dominate the biomedical VLM landscape, particularly in scenarios that require text generation grounded in visual understanding. This prevalence naturally motivates the exploration of alternative generative paradigms.
\begin{figure}[ht]
  \centering
  \includegraphics[width=0.9\linewidth]{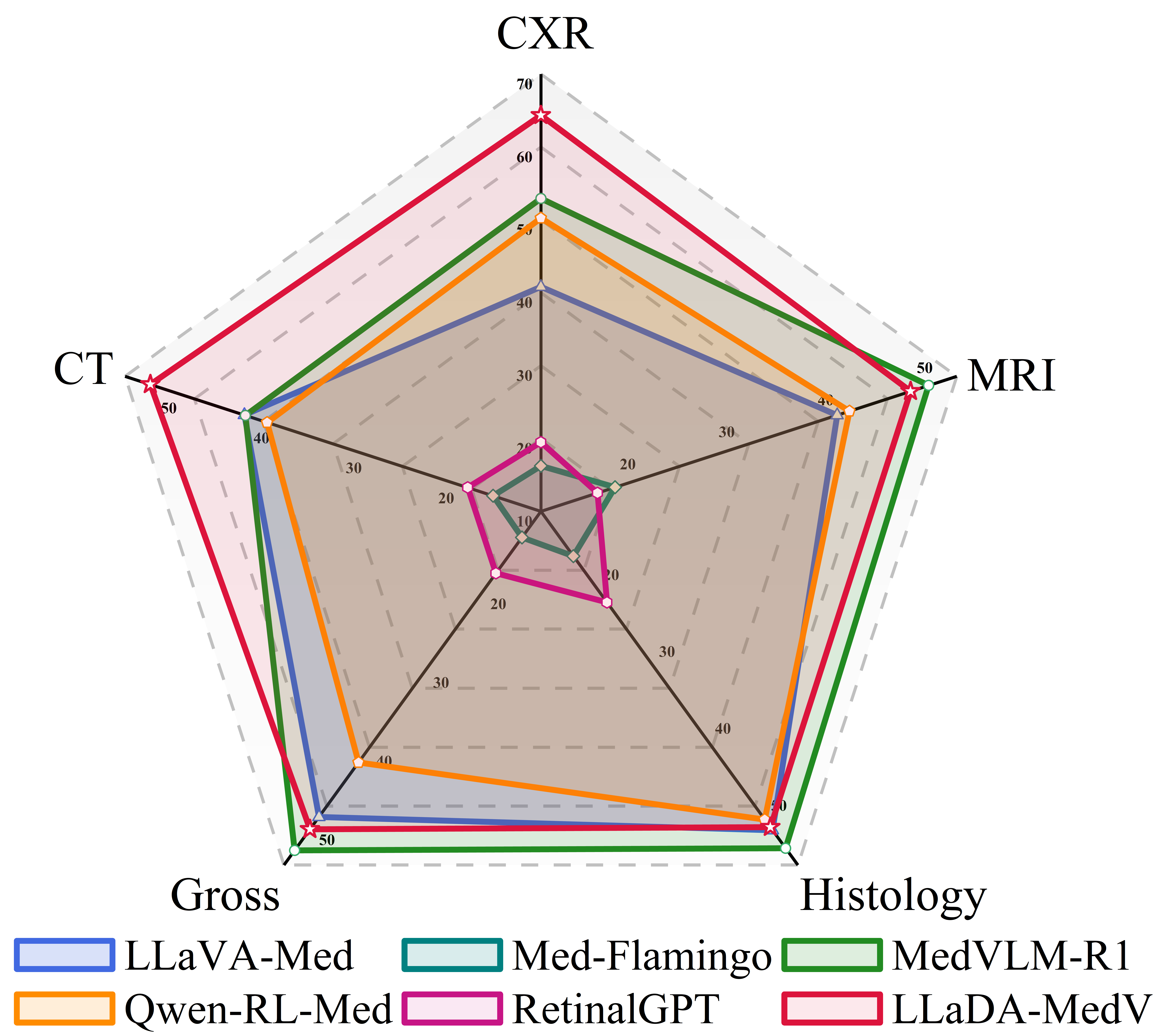}
  \caption{Illustration of biomedical VLMs evaluated in the open-ended biomedical conversation benchmark. Among the 6 Medical VLMs, LLaDA-MedV achieves the highest overall score and demonstrates the best performance on Chest X-ray (CXR) and CT modalities.}
  \vspace{-0.4cm}
  \label{fig:radar}
\end{figure}
Recently, masked diffusion models (MDMs)~\cite{lou2023discrete,shi2024simplified,sahoo2024simple} have shown significant promise in language generation. In particular, LLaDA~\cite{nie2025large} have emerged as a compelling alternative. Unlike continuous-state diffusion models~\cite{ho2020denoising,song2020denoising,song2020score}, which operate over continuous data (e.g., image pixels), it operate directly on discrete tokens. In the forward process, it replaces input tokens with a special \texttt{\textless mask\textgreater} tokens, which acts as an absorbing state. In the reverse generation process, the model predicts the masked content simultaneously, progressively refining the sequence across multiple steps. Notably, with large-scale training, LLaDA-based models~\cite{nie2025large,you2025llada} have already demonstrated strong scalability and competitive performance in the general domain.

Although language diffusion models have shown strong performance in general domains, applying them to biomedical image understanding remains nontrivial. The core difficulty lies in the substantial domain gap between general and biomedical data. To move forward, we must first address several key questions critical to building effective biomedical diffusion VLMs:
\begin{itemize}
\item \textit{How can the success of language diffusion models in the general domain be effectively adapted to biomedical image understanding?}
\item \textit{Why are language diffusion models promising for biomedical vision-language modeling?}
\item \textit{What design principles are essential for developing effective biomedical diffusion VLMs?}
\end{itemize}

To address these questions, we first introduce \textbf{LLaDA-MedV}, a large language diffusion model specifically developed for biomedical visual understanding. LLaDA-MedV demonstrates strong visual understanding ability. As shown in Fig.~\ref{fig:radar}, the model outperforms several autoregressive baselines on open-ended biomedical visual conversation tasks, following semantic alignment and supervised fine-tuning. 
Second, we conduct a comparative analysis between LLaDA-MedV and LLaVA-Med~\cite{li2023llava}. Our results show that LLaDA-MedV produces significantly longer and more detailed responses, while also enabling explicit control over output length, which is an important feature for tasks that demand comprehensive answers.
Finally, we further perform an in-depth analysis of the training and inference pipelines to identify the key factors underlying the model performance. On the training setting, we find that proper initialization and domain-specific fine-tuning are critical. During inference, we analyze the role of sampling steps and observe that they substantially impact output diversity, especially in longer sequence generation.

In summary, our key contributions are as follows:

\begin{itemize}
    \item We present \textbf{LLaDA-MedV}, the first diffusion-based vision-language model designed for biomedical image understanding via visual instruction tuning.
    
    \item We provide a comprehensive empirical study comparing LLaDA-MedV with autoregressive counterparts across open-ended biomedical conversations and VQA tasks, showing consistent advantages in response quality and controllability.
    
    \item We conduct in-depth analysis of training and inference behavior of LLaDA-MedV, identifying several design factors that affect generation performance (e.g., sampling and initialization). These findings offer valuable insights for future research on diffusion-based language models, particularly in biomedical field.
\end{itemize}

\section{Related Works}
\subsection{Large Language Diffusion Models} Recent advancements in diffusion models for vision tasks have spurred growing interest in their application to natural language processing. However, adapting diffusion models to language presents unique challenges due to the discrete nature of language tokens, which contrasts with the continuous pixel representations in vision tasks. To address this, several studies have proposed learning continuous representations of textual data~\cite{gong2022diffuseq,strudel2022self,chen2022analog,richemond2022categorical,mahabadi2023tess,ye2023dinoiser}. Moreover, Diffusion-LM~\cite{li2022diffusion}  introduces a method to denoise sequences of Gaussian vectors into word embeddings using learned bidirectional mappings. 
In parallel, other works focus on discrete diffusion frameworks, particularly masked diffusion models (MDMs)~\cite{ou2024your,gong2024scaling}. Notably, LLaDA~\cite{nie2025large} marks a significant milestone by scaling MDMs to 8B parameters and achieving competitive results compared to ARM baselines such as LLaMA3~\cite{grattafiori2024llama}.

Despite these promising developments, most diffusion-based language models are restricted to text-only applications. Their potential in multimodal contexts, especially in the biomedical domain, remains largely unexplored. This gap forms the basis for our work, which aims to extend the benefits of language diffusion models to the biomedical domain.

\subsection{BioMedical VLMs} Biomedical VLMs typically leverage existing general-domain architectures either by training from scratch or through domain-specific fine-tuning~\cite{moor2023med,pan2025medvlm,lai2025med}. For instance, BiomedGPT~\cite{luo2023biomedgpt,zhang2023biomedgpt} adopts a GPT-style autoregressive architecture and is trained from scratch on large-scale biomedical multimodal datasets, achieving strong performance across several biomedical benchmarks. In addition to large-scale training strategies that demand substantial computational resources, prompt learning and prompt engineering have emerged as efficient alternatives~\cite{wang2024does,denner2024visual,guo2024prompting}. For example, ChatCAD~\cite{wang2024interactive} integrates large language models into computer-aided diagnosis systems, generating clinically grounded outputs to support decision-making. 

While these method have advanced the field significantly, most biomedical VLMs to date rely on autoregressive modeling. The potential of discrete diffusion models, particularly MDMs, remains largely untapped in biomedical applications. To address this gap, we propose LLaDA-MedV, the first biomedical VLM based on masked diffusion modeling. Furthermore, our work opens a new direction for scalable, effective biomedical image-language understanding using diffusion based LLM model.

\section{Preliminary}\label{sec:preliminary}

\subsection{Masked Diffusion Models} 
Masked diffusion models (MDMs) operate over discrete tokens (e.g., text) using a structured forward–reverse process defined over a finite support. In LLaDA, a special mask token $\texttt{\textless mask\textgreater} :=\mathbf{M}$ is introduced, which serves as an absorbing state in the transition matrix. Intuitively, as the time step $t$ increases, each token either remains unchanged or transitions to $\mathbf{M}$ with probability $t$~\cite{austin2021structured,nie2025large}. This forward process can be formally described as:
\begin{equation}\label{eq:forward-llada-main}
    q_{t|0}(x_t \mid x_0) = \prod_{i=1}^{L} q_{t|0}(x_t^i \mid x_0^i)
\end{equation}

\begin{equation}
    \text{where} \quad
    q_{t|0}(x_t^i \mid x_0^i) =
    \begin{cases}
        1 - t, &  x_t^i = x_0^i \\
        t,     &  x_t^i = \mathbf{M}
    \end{cases}\nonumber
\end{equation}
Thus, as $t$ increases, a greater proportion of tokens are replaced by the mask token $\mathbf{M}$. Analogous to denoising score matching in continuous diffusion models~\cite{vincent2011connection}, the reverse process in LLaDA seeks to reconstruct the original sequence from a fully masked sequence. Specifically, for a reverse step where $0 \leq s < t \leq 1$, the conditional transition is defined as:
\begin{equation}\label{eq:backward-llada-main}
    q_{s|t}(x_s|x_t)=\prod_{i=1}^L q_{s|t}(x_s^i\mid x_t) \quad \text{where}
\end{equation}

\begin{equation}
    q_{s|t}(x_s^i \mid x_t) =
    \begin{cases}
        1 , &  x_t^i \ne \mathbf{M},  x_s^i = x_t^i \\
        \frac{s}{t},      &x_t^i = \mathbf{M}, x_s^i = \mathbf{M} \\
        (1-\frac{s}{t}) q_{0|t}(x_s^i\mid x_t) & x_t^i = \mathbf{M}, x_s^i \ne \mathbf{M} \\
        0  &\text{otherwise}
    \end{cases}\nonumber
\end{equation}
This formulation intuitively captures the behavior of the reverse process: it retains the unmasked tokens, maintains a portion $s/t$ of the masked tokens as masks, and samples the remaining $(1 - s/t)$ proportion from $q_{0|t}(\cdot)$. \cite{ou2024your} propose an efficient approach to compute $q_{0|t}(\cdot)$ based on clean data prediction, such that $q_{0|t}(x_s^i\mid x_t) = p_d(x_0^i \mid x_t)$ for all positions where $x_t^i = \mathbf{M}$. To model the unknown data distribution $p_d(\cdot)$, a mask predictor $p_\theta(\cdot|x_t)$ is introduced. The learning objective follows the formulations proposed in~\cite{lou2023discrete,nie2025large}:
\begin{equation}\label{eq:lladaMain}
    \mathcal{L}_\theta^0 := -\mathbb{E}_{t, x_0, x_t}[\frac{1}{t}\sum_{i=1}^{L}\textbf{1}[x_t^i=\mathbf{M}]\log p_\theta(x_0^i|x_t)]
\end{equation}
Here, $\textbf{1}[\cdot]$ is the indicator function and $t \sim \mathcal{U}[0, 1]$. 

At inference time, generation proceeds as motivated by Eq.~\ref{eq:backward-llada-main}. Beginning with a fully masked sequence (i.e., $t = 1$), the model uses learned $p_\theta(x_0 \mid x_t)$ to iteratively reconstruct the clean sequence. A remasking step is applied at each stage to ensure alignment with the theoretical reverse dynamics (i.e., retain $s/t$ portion masked components). This denoising process continues until the final step, resulting in a fully reconstructed output.

\subsection{Visual Intruction Tuning} Rather than training VLMs from scratch, visual instruction tuning~\cite{liu2023visual} (i.e., LLaVA) offers a more efficient paradigm for equipping a pretrained language backbone with visual understanding. This approach adopts a modular architecture comprising three key components: a language backbone $f_\phi(\cdot)$ (e.g., LLaMA~\cite{touvron2023llama}), a vision encoder $g(\cdot)$ (e.g., CLIP~\cite{radford2021learning}), and a projection module $h(\cdot)$ (typically a lightweight MLP). Given a visual input $X_v$, the vision encoder first extracts high-level image features $g(X_v)$, which are then projected into the language embedding space via $h(\cdot)$ to produce $H_v = h(g(X_v))$. These vision embeddings are inserted into the language model's input sequence, often prepended to the text prompt, and jointly processed by $f_\phi$. This setup allows the language model to generate coherent, visually grounded responses conditioned on both image and text inputs, thereby achieving multimodal capabilities with minimal architectural modifications. To support effective training, the input data are typically formatted as single/multi-turn dialogues that simulate interactions between a human user and an AI assistant~\cite{liu2023visual,li2023llava,liu2024improved}. 

\section{Methods}\label{sec:method}
\subsection{Training Architecture} 
\textbf{Learning objective}. For clarity, we describe the setup using a single image and a single-turn dialogue. Given training instance denoted by a tuple $(X_v, u_0, r_0)$, where $X_v$ represent the image, $u_0 := [u_0^i]_{i=1}^{L_{u_0}}$ denote the user prompt (e.g., a biomedical query) of length $L_{u_0}$ and $r_0:= [r_0^j]_{j=1}^{L_{r_0}}$ represents the ground-truth assistant response (e.g., generated by GPT) of length $L_{r_0}$.The training objective extends the original masked prediction formulation (i.e., Eq.~\ref{eq:lladaMain}) by conditioning on the user prompt and is defined as:
\begin{equation}\label{eq:lladaSFT}
    \mathcal{L}_\theta^1 := -\mathbb{E}[\frac{1}{t}\sum_{j=1}^{L_{r_0}}\textbf{1}[r_t^j=\mathbf{M}]\log p_\theta(r_0^j|X_v, u_0,r_t)]
\end{equation}
Here $r_t$ denotes the masked response obtained from the clean response $r_0$ through forward masking, and we abuse notation by letting $X_v$ also represent the corresponding visual embeddings for brevity. Intuitively, this objective trains the mask predictor $p_\theta$ to recover the masked portions of the assistant's response based on both the user prompt and visual features. Following prior works~\cite{nie2025large,you2025llada}, we model $p_\theta$ by a Transformer architecture with bidirectional attention.

\textbf{Multi-stage Training Pipeline}. Training LLaDA-MedV consists of three stages. The first two stages aim to establish semantic alignment between biomedical language and visual content, and to equip the model with instruction-following capabilities for biomedical visual understanding. To further improve service quality in data-specific scenario, we perform an additional SFT stage using three biomedical VQA training datasets. Details of each stage are provided below:
\begin{itemize}
    \item \textbf{Stage 1: Biomedical Semantic Alignment}. In this stage, we freeze both the vision tower and the language backbone, and fine-tune only the lightweight MLP projector. This step ensures that the extracted visual features are effectively projected into the language embedding space and semantically aligned with biomedical concepts. 
    \item \textbf{Stage 2: End-to-End Vision Instruction Tuning}. After stage 1, we fine-tune the language backbone and the projector module to enable LLaDA-MedV the medical visual understanding and coherent response generation abilities. Unlike~\cite{you2025llada}, we keep the vision tower frozen. Each training instance consists of a single image and its associated multi-turn dialogues.
    \item \textbf{Stage 3: Dataset Specific Fine-tuning}. To further improve model performance in scenario that require higher accuracy, we further fine-tune the model on three benchmarks~\cite{lau2018dataset,liu2021slake,he2020pathvqa}. Each training example is formatted as a single-turn dialogue between a human user and the assistant, following the same training setup as in Stage 2. This step enables LLaDA-MedV to provide free-form answers for both closed-form and open-ended biomedical questions. The vision tower remains frozen during this stage as well.
\end{itemize}
It is important to note that we do not adopt the initialization strategy used in~\cite{li2023llava}. In our experiments, initializing from weights~\cite{you2025llada} degraded the model’s ability to interpret medical images and led to repetitive output generation. \textit{\textbf{For implementation details, please refer to the Appendix~A}}.

\subsection{Inference Architecture}
Generating responses simulates the reverse dynamics (as defined in Eq.~\ref{eq:backward-llada-main}). Beginning with a fully masked response, the learned mask predictor $p_\theta$ progressively reconstructs the response content, accompanied by appropriate remasking to align with the reverse diffusion process. Following prior works~\cite{you2025llada,nie2025large}, we adopt low-confidence remasking strategy~\cite{chang2022maskgit}, where only tokens with low confidence (i.e., lower $p_\theta$ values) are remasked. In addition, we also explore a semi-autoregressive generation strategy~\cite{nie2025large}. Specifically, a response of length $L$ is divided into $L/B$ blocks, where $B$ denotes the block length. The generation process is applied sequentially from left to right across these blocks, with each block undergoing $ Z\cdot B /L$ sampling steps. \textit{\textbf{We provide the detailed inference algorithm in Appendix~A}}.

\begin{table*}[ht]
\centering
\resizebox{0.95\textwidth}{!}{
\begin{tabular}{l|cc|ccccc|c}
\midrule
& \multicolumn{2}{c|}{\textbf{Question Types}} & \multicolumn{5}{c|}{\textbf{Domains}} & \textbf{Overall} \\
& Conversation & Description & CXR & MRI & Histology & Gross & CT & \\
& (143) & (50) & (37) & (38) & (44) & (34) & (40) & (193) \\
\midrule
LLaMA~\cite{meta2024llama3.2-11b-vision} & 27.850 & 27.750 & 33.505& 23.841& 24.684& 31.764 & 26.458 & 27.824\\
LLaVA~\cite{liu2023visual} & 43.509 & 29.671 & 45.935 & 41.761 & 42.726 & 33.905 & 33.4 & 34.653 \\
LLaVA-Med~\cite{li2023llava} & 51.750 & 24.730 & 40.819 & 39.928 & 51.452 & 48.880 & 42.083 & 44.750 \\
Med-Flamingo~\cite{moor2023med} & 16.339 & 13.702 & 16.238 & 17.502 & 15.823 & 13.311 & 15.174 & 15.656 \\
MedVLM-R1~\cite{pan2025medvlm} & 52.774 & 42.663 & 52.944 &\textbf{49.159} & 53.801 & \textbf{53.122} & 41.984 & 50.154 \\
Qwen-RL-Med~\cite{zhu2025toward} & 46.408 & 40.016 & 50.225 & 41.181 & 50.081 & 41.929 & 39.618 & 44.752 \\
Qwen-VL~\cite{Qwen-VL}& 51.546 & 40.143 & 58.929 & 43.186 & 51.344 & 47.106 & 42.401 & 48.592 \\
RetinalGPT~\cite{zhu2025retinalgpt} & 20.535 & 13.349 & 19.477 & 15.722 & 21.839 & 17.875 & 17.932 & 18.673 \\
\midrule
LLaDA-V~\cite{you2025llada} & 53.574 & 42.627 & 57.545 & 38.445 & \textbf{64.191} & 49.055 & 42.753 & 50.738 \\
\cellcolor[gray]{0.85}Ours & \cellcolor[gray]{0.85}\textbf{54.141} & \cellcolor[gray]{0.85}\textbf{48.214} & \cellcolor[gray]{0.85}\textbf{64.352} & \cellcolor[gray]{0.85}47.348 & \cellcolor[gray]{0.85}51.073 & \cellcolor[gray]{0.85}50.432 & \cellcolor[gray]{0.85}\textbf{50.268} & \cellcolor[gray]{0.85}\textbf{52.605} \\
\midrule
\end{tabular}}
\caption{Illustration of results on the open-ended biomedical conversation benchmark across several VLM baselines. CXR refers to Chest X-ray. Scores reflect relative performance against GPT-4 responses, as judged by GPT-4.1. For models with intermediate reasoning steps, we include all content for fair comparison.}
\label{tab:chatbot}
\vspace{-0.4cm}
\end{table*}

\section{Experiments}
\subsection{Experimental Setup and Configuration}
\textbf{Dataset and Configuration.} Training LLaDA-MedV involves three stages. In Stage 1, we use 600k alignment text-image pairs. In Stage 2, we employ a 60k multi-turn inline-mention dialogue dataset, selected for its consistently strong performance. In Stage 3, we further fine-tune the model using the training sets from VQA-RAD~\cite{lau2018dataset}, SLAKE~\cite{liu2021slake}, and PathVQA~\cite{he2020pathvqa}, following the official data splits provided in~\cite{li2023llava}.
For the model architecture, we use LLaDA-8B-Instruct~\cite{nie2025large} as the language backbone. The vision tower is based on SigLIP2~\cite{tschannen2025siglip}, and visual-language projector is implemented as a lightweight two-layer MLP with GELU activation. All training is performed on four NVIDIA A100 80GB GPUs. 

For evaluation, we set the hyperparameters to $L=256$, $B=64$, and $Z=256$ for open-ended biomedical conversation evaluation. To ensure fair comparison, we also set the maximum token length to 256 for all baseline models. For the three downstream biomedical VQA tasks, we use $L = B = Z = 64$ to ensure computational efficiency, and directly report the official scores of the baseline models as reported in~\cite{li2023llava}. Unless otherwise specified, these parameter settings are used consistently across all experiments. \textit{\textbf{Additional details are provided in the Appendix~A}}.

\begin{figure}[t]
  \centering
  \includegraphics[width=0.9\linewidth]{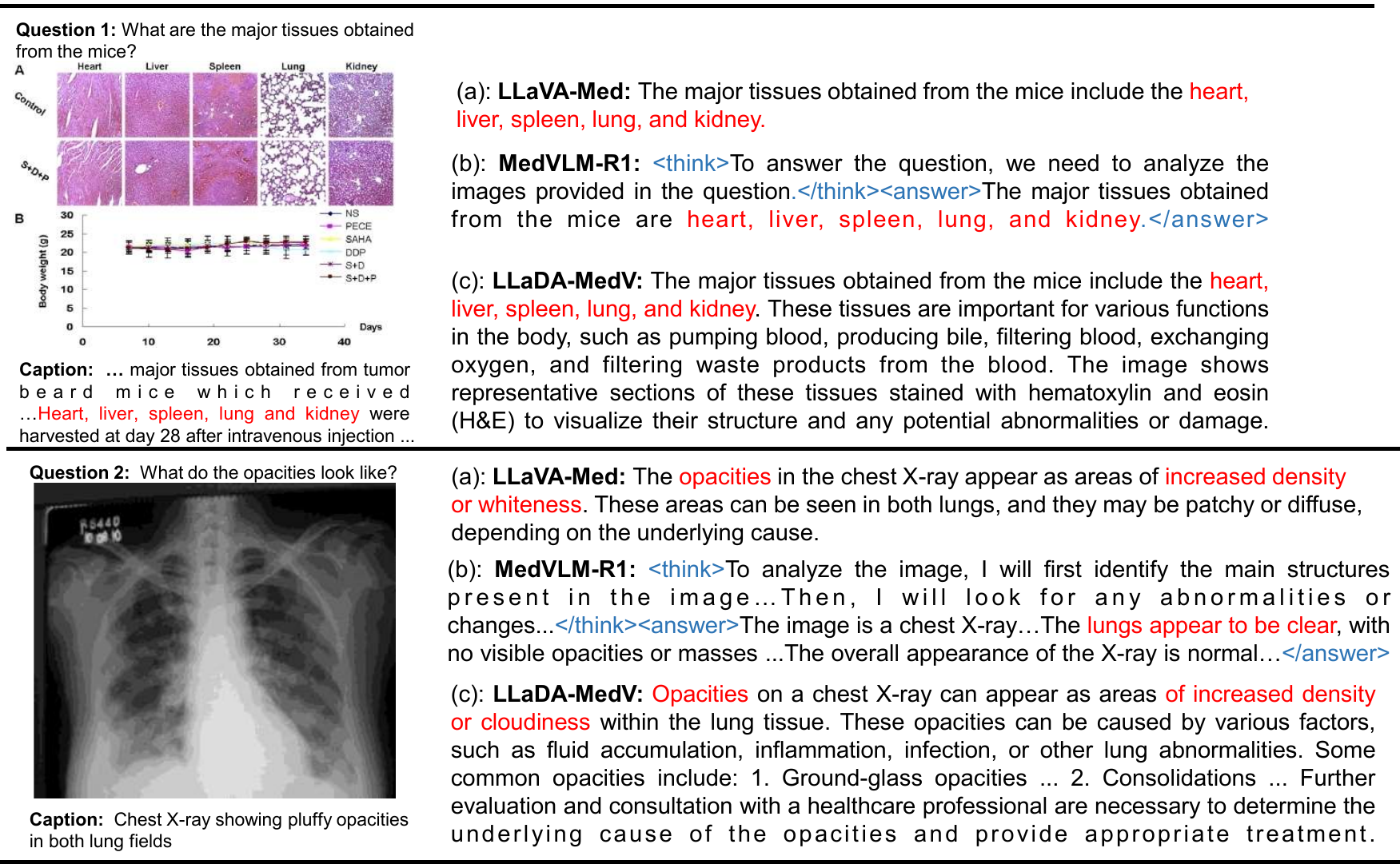}
  \caption{Illustration of open-end conversation evaluation. All questions, images and corresponding captions are sourced from~\cite{li2023llava}. We present representative responses from (a) LLaVA-Med, (b) MedVLM-R1, and (c) LLaDA-MedV. Key informative segments are highlighted in red for emphasis.}
  \vspace{-0.4cm}
  \label{fig:chatbot}
\end{figure}

\subsection{Open-end Biomedical Conversation}

We evaluate LLaDA-MedV on the Biomedical Visual Chatbot benchmark~\cite{li2023llava}, which measures performance on open-ended, image-grounded questions. Specifically, given the ground-truth figure captions and questions, an LLM judge compares the model-generated response (e.g., LLaDA-MedV) with a fixed reference answer produced by GPT-4~\cite{openai2023gpt4}. The judge assigns each answer a score from 1 to 10 based on helpfulness, relevance, accuracy, and level of detail. Scores are then averaged within each question type (e.g., conversational) and domain (e.g., MRI), and the resulting averages are subsequently used to compute a score relative to the reference, which serves as our primary evaluation metric.

Two points need clarifications. First, the GPT-4 reference answers are fixed and provided by the benchmark~\cite{li2023llava} and the LLM judge evaluates model outputs only against these references. Second, we use GPT-4.1 mini~\cite{openai2025gpt41mini} as the judging model because the original GPT-4-0314 used in~\cite{li2023llava} is no longer accessible. To ensure fairness, we therefore re-evaluate all baseline systems rather than relying on their originally reported scores.

LLaDA-MedV demonstrates superior ability in following visual instructions and generating informative responses. As shown in Tab.~\ref{tab:chatbot}, it outperforms both LLaVA-Med and MedVLM-R1 across nearly all subjects, achieving overall performance gains of 7.855\% and 2.45\%, respectively.  Furthermore, we observe that LLaDA-MedV tends to produce longer responses compared to ARM baselines. For instance, as illustrated in Question 2 of Fig.~\ref{fig:chatbot}, instead of merely describing the appearance of opacities, LLaDA-MedV also provides relevant contextual information such as potential causes, common categories, and a cautious recommendation for further consultation. \textit{\textbf{We provide more case studies in Appendix~B.}}

Two additional points warrant clarification. First, unlike other ARM baselines (e.g., LLaVA-Med over LLaMA), our language backbone (i.e., LLaDA) is trained solely via pretraining and SFT, without any post-training such as reinforcement learning (RL)~\cite{guo2025deepseek} or human preference tuning~\cite{ouyang2022training}. Second, although AI assistants such as GPT are not perfect evaluators, their use has become standard practice in the community~\cite{liu2023g,hsu2023gpt,yang2024gpt,li2023llava}. Therefore, we believe our comparative evaluation remains fair and sufficiently rigorous to demonstrate the advantages of our approach in biomedical field.
\begin{table*}[h]
\centering
\resizebox{0.9\textwidth}{!}{
\begin{tabular}{l|cc|cc|cc}
    \midrule
& \multicolumn{2}{c|}{\textbf{VQA-RAD}} & \multicolumn{2}{c|}{\textbf{SLAKE}} & \multicolumn{2}{c}{\textbf{PathVQA}}   \\
\textbf{Model}&  Open & Closed &  Open & Closed &  Open & Closed \\
     \midrule
LLaVA & 50.00  & 65.07 & 78.18 & 63.22 & 7.74 & 63.20 \\
LLaVA-Med (From LLaVA) & 61.52 & 84.19 & 83.08 & 85.34 & 37.95 &91.21 \\
VL Encoder-Decoder~\cite{bazi2023vision} &  71.49 & 82.47 & -  &-  & \textbf{71.49} & 85.61 \\
Q2ATransformer~\cite{liu2023q2atransformer} & \textbf{79.19}& 82.47 &-  & - & 54.85& 88.85 \\
Prefix T. Medical LM~\cite{van2023open} & - & - & \textbf{84.30} & 82.01 & 40.00 & 87.00 \\
PubMedCLIP~\cite{eslami2023pubmedclip} & 60.10 & 80.00 & 78.40 & 82.50 & - & - \\
BiomedCLIP~\cite{zhang2023large} & 67.60 & 79.80 & 82.05 & 89.70 & - & - \\
M2I2~\cite{li2023self} & 66.50 & 83.50 & 74.70 & 91.10 & 36.30& 88.00 \\
    \midrule
\cellcolor[gray]{0.85} Ours & \cellcolor[gray]{0.85} 45.60 & \cellcolor[gray]{0.85} \textbf{84.93}& \cellcolor[gray]{0.85}68.85 & \cellcolor[gray]{0.85}\textbf{92.31}& \cellcolor[gray]{0.85}31.96 & \cellcolor[gray]{0.85}\textbf{95.15} \\
    \midrule
\end{tabular}
}
\caption{
Illustration of results on downstream VQA tasks. For all baselines, we report results directly when available. Following~\cite{li2023llava}, we use token-level recall for open-form questions and answer accuracy for closed-form ones. Prior models often treat open-form questions as classification over a fixed answer set, raising concerns about generalizability. In contrast, due to limited post-training, LLaDA-MedV may struggle to perform reliable classification over predefined candidate sets, instead producing more open-ended responses that are harder to constrain within rigid answer formats.}
\label{tab:downstreamqa}
\end{table*}

\begin{figure}[ht]
  \centering
  \includegraphics[width=0.95\linewidth]{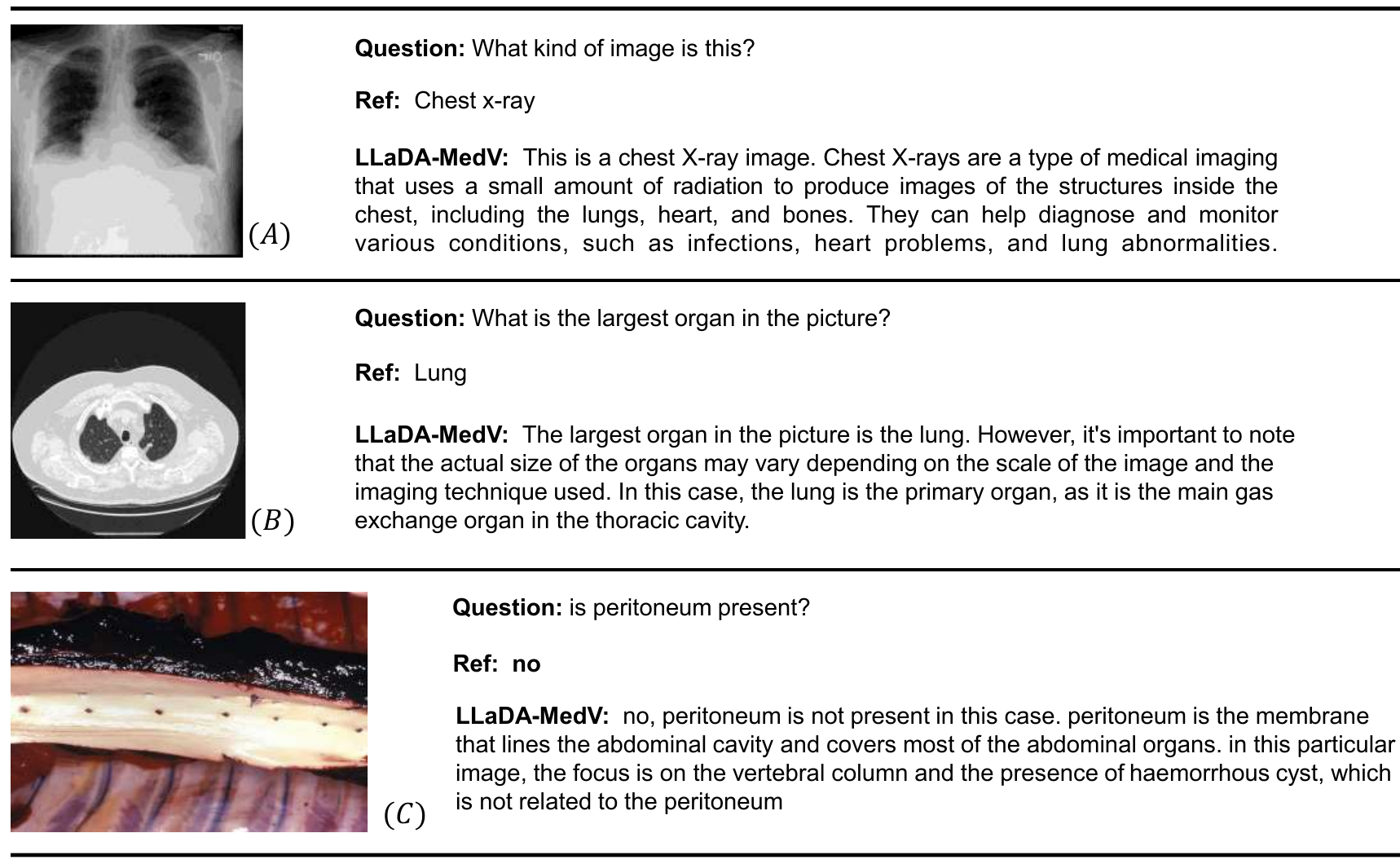}
  \caption{Illustration of LLaDA-MedV responses on downstream VQA tasks across three benchmarks. \textbf{(A)} and \textbf{(B)} represent the open-form QA examples from VQA-RAD and SLAKE benchmarks, respectively. \textbf{(C)} represents the closed-form (i.e., yes/no) QA from PathVQA benchmark. For this experiment, we set $L = B = Z = 64$.
  }
  \label{fig:vqa}
\end{figure}

\subsection{Downstream Visual Question Answering}
We evaluate model performance on three biomedical visual question answering (VQA) benchmarks, which are designed to assess the model’s accuracy in clinically relevant scenarios. We follow the evaluation strategy proposed in~\cite{li2023llava}. For closed-form questions (e.g., Yes/No), we report accuracy. For open-form questions, we report recall, measuring the proportion of ground-truth tokens present in the generated sequence.

LLaDA-MedV demonstrates competitive performance compared to ARM baselines. As shown in Tab.~\ref{tab:downstreamqa}, our model achieves the highest accuracy on closed-form questions across all three benchmarks, with scores of 84.93\% on VQA-RAD, 92.31\% on SLAKE, and 95.15\% on PathVQA. Additionally, as shown in Fig.~\ref{fig:vqa}, the model’s answers are not only correct but also more informative. For example, in Fig.~\ref{fig:vqa}(C), our model recognizes that the peritoneum is absent from the image and provides a specific justification (e.g., the image focuses on the vertebral column).  However, performance on open-form questions remains less competitive. We observe that answering open-form questions in the VQA setting is inherently more challenging for our model, as the underlying language backbone (i.e., LLaDA) lacks sufficient post-training. Consequently, LLaDA-MedV struggles to model open-form questions as a classification task over a predefined answer set derived from the training split, which is the strategy commonly adopted in prior works~\cite{li2023llava}. 
Despite this limitation, the results underscore LLaDA-MedV's potential in dataset-specific biomedical applications, motivating future efforts to improve its instruction-following and answer-formatting capabilities through post-training (e.g., reinforcement learning from human feedback~\cite{ouyang2022training}). 
\begin{figure}[t]
  \centering
  \includegraphics[width=1\linewidth]{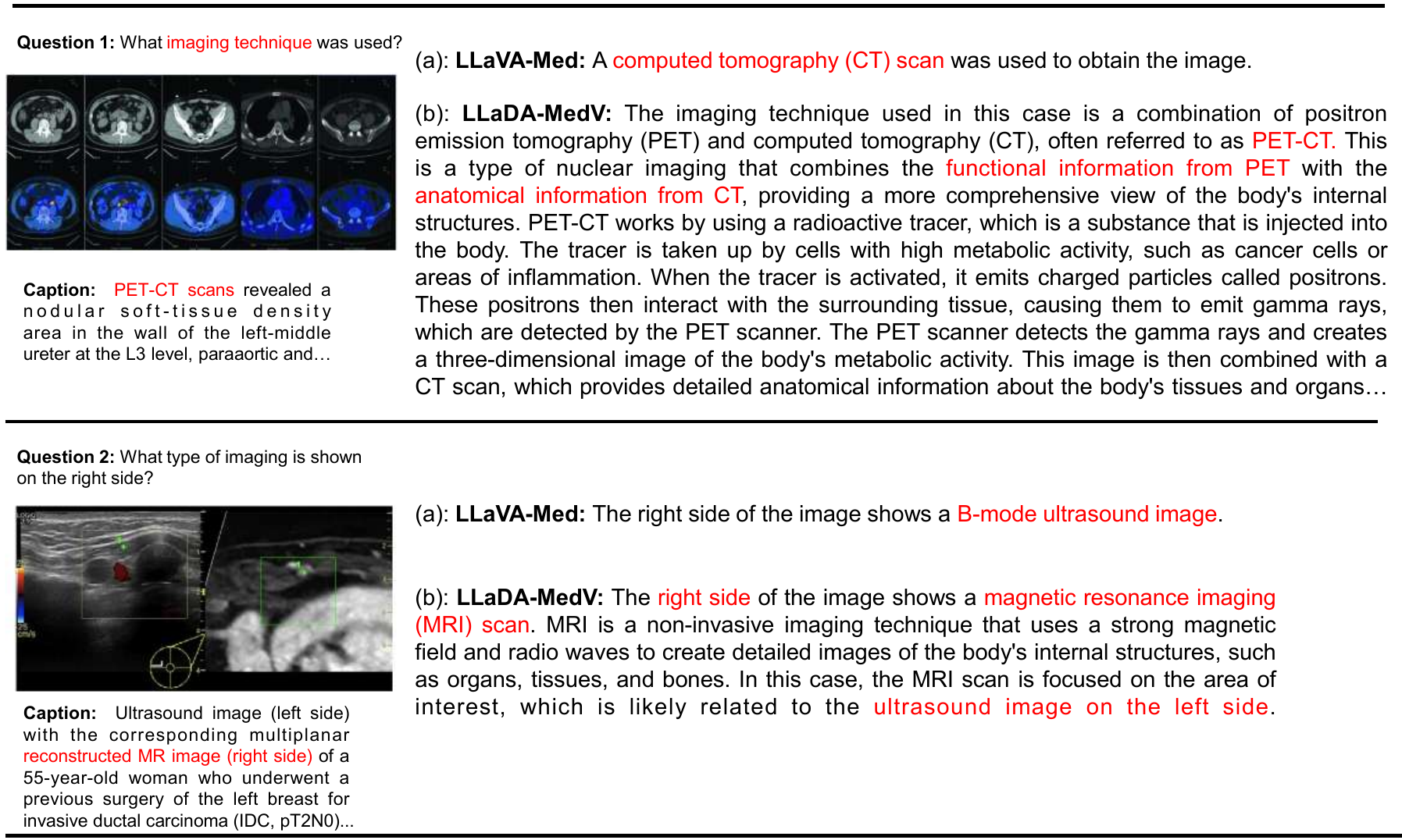}
  \caption{Illustration of LLaVA-Med and LLaDA-MedV responses to biomedical queries 1 and 2. The images, queries, and corresponding captions are adapted from~\cite{li2023llava}. For fair comparison, the generation lengths are aligned by setting the maximum token limit to the same value (e.g., generation length $L = 256$).}

  \label{fig:compare-figure2}
\end{figure}

\begin{table}[ht]
    \centering
    \resizebox{1\columnwidth}{!}{%
    \begin{tabular}{l|c|c|c|c|c}
        \midrule
        \textbf{OE Conversation (193)} & \textbf{T/Q} & \textbf{W/Q} & \textbf{T/W} &\textbf{Z} & \textbf{Overall Score} \\
        \midrule
        LLaVA-Med & 1.317 & 36.332 & 0.036 & - & 44.750 \\
        LLaVA-Med$^{200}$ & 1.392 & 40.922 & 0.034 & - & 44.582  \\
        \midrule
        LLaDA-MedV$_{256}$ & 38.382 & 166.585 & 0.230 & 256 & 52.605 \\
        LLaDA-MedV$_{128}$ & 30.874&170.399 &0.181 & 128 & 44.276 \\
        LLaDA-MedV$_{64}$ & 21.234 & 172.839 & 0.123 & 64 & 28.523 \\
        LLaDA-MedV$_{32}$ & 13.945 & 182.596 & 0.076 & 32 & 18.581 \\
        LLaDA-MedV$_{16}$ & 8.595 & 192.09 & 0.045 & 16 & 13.525 \\
        \midrule
    \end{tabular}}
    \caption{Illustration of open-ended (OE) conversation evaluation on 193 questions. \textbf{T/Q} is the average response time per question (in seconds), \textbf{W/Q} is the average word count, and \textbf{T/W} is the average time per word. LLaVA-Med$^{200}$ uses a modified prompt encouraging responses of at least 200 words. For LLaDA-MedV, we fix the generation length $L=256$ and block size $B=64$, and vary the sampling steps $Z$ from 16 to 256 during inference.}  
    \label{tab:response-tw}
\end{table}

\section{Analysis}

\subsection{Why Language Diffusion Models?}

ARMs typically control response length through heuristics such as adjusting the maximum token limit or modifying the system prompt. However, this method often proves unreliable in biomedical contexts. For example, models like LLaVA-Med frequently terminate early when an end-of-sequence token is predicted prematurely, resulting in responses that are shorter than expected and lacking in detail. As shown in Tab.~\ref{tab:response-tw}, LLaDA-MedV generates an average of 166 words per response, whereas LLaVA-Med generates only around 36. To examine whether prompt engineering could extend response length in LLaVA-Med, we introduced a system prompt explicitly requesting at least 200 words per answer. Despite this, the model's output increased only marginally, reaching approximately 40 words per response. These findings suggest that ARMs offer limited control over output length in practice.

In contrast, LLaDA-MedV follows a different generation process. It begins with a fully masked sequence of fixed length and gradually fills in the tokens through iterative denoising. This design enables the model to explicitly control the length of the generated response, resulting in outputs that are both longer and more complete. As illustrated in Fig.~\ref{fig:compare-figure2}, the model not only identifies the imaging modality correctly (e.g., PET/CT) but also explains how PET is combined with CT, providing helpful contextual insights absent from ARM-based outputs. This increase in content is reflected in automatic evaluation scores. As reported in Tab.~\ref{tab:chatbot}, LLaDA-MedV achieves a 7.855\% improvement in response quality compared to LLaVA-Med. LLaDA-MedV requires more computation. As shown in Tab.~\ref{tab:response-tw}, it raises inference time per word from 0.036 seconds to 0.230 seconds. However this trade-off is reasonable given the improved output quality. Moreover, the current implementation of LLaDA-MedV is not yet optimized, and we anticipate significant gains in efficiency with further engineering. Overall, these results demonstrate that masked language diffusion offers more reliable control over response length and content, which is particularly valuable for biomedical applications that demand detailed and context-rich answers.

\begin{table}[ht]
    \centering
    \resizebox{\columnwidth}{!}{%
    \begin{tabular}{l|c|cc|cc}
        \midrule
         & \textbf{OE Conversation} & \multicolumn{2}{c|}{\textbf{VQA-RAD}} & \multicolumn{2}{c}{\textbf{SLAKE}} \\
        \multicolumn{1}{c|}{\textbf{Model}} & Overall Score & Open & Ended & Open & Ended \\
        \midrule
        LLaVA-Med$^*$ & 44.750 & 28.23 & 61.40 & 39.17 & 52.16 \\
        LLaDA-MedV$^*$ & 52.605 & 39.34 & 75 & 45.96 & 67.31 \\
        \midrule
        LLaDA-MedV$^{V_1}$ & 31.123 & - & - & - & - \\
        LLaDA-MedV$^{V_2}$ & 31.056 & - & - & - & - \\
        \midrule
    \end{tabular}}
    \caption{Illustration of performance in analyzing the training pipeline. LLaVA-Med$^*$ and LLaDA-MedV$^*$ are trained with only alignment and SFT. LLaDA-MedV$^{V_1}$ uses LLaDA-V initialization and is trained with SFT only. LLaDA-MedV$^{V_2}$ also starts from LLaDA-V but includes both alignment and SFT. All LLaDA-based models use identical training and inference settings as main experiments. Due to the large performance gap, no further fine-tuning were performed on LLaDA-MedV$^{V_1}$ and LLaDA-MedV$^{V_2}$.}  
    \label{tab:compare-training}
\end{table}

\subsection{Training-Time Considerations}

Task-specific fine-tuning plays an important role in optimizing performance for specialized biomedical benchmarks. As shown in Tab.~\ref{tab:compare-training}, although LLaDA-MedV outperforms LLaVA-Med in zero-shot settings on the VQA-RAD and SLAKE benchmarks, a clear performance gap remains when compared to fine-tuned models (Tab.~\ref{tab:downstreamqa}). These findings suggest that fine-tuning can provide substantial benefits in improving performance and enhancing assistant capabilities in biomedical scenarios.

\begin{figure}[t]
  \centering
  \includegraphics[width=\linewidth]{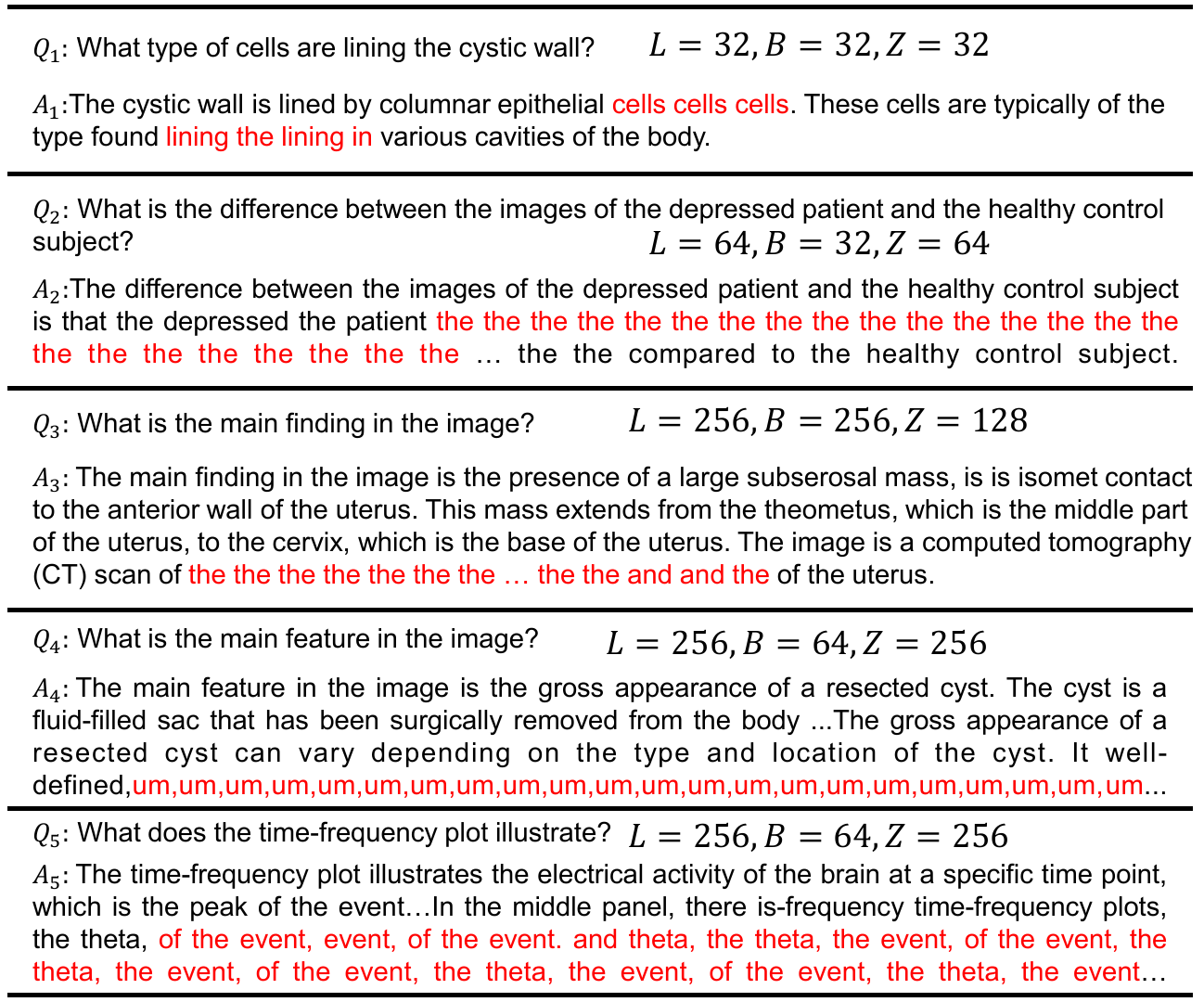}
  \caption{Illustration of token repetition during generation (i.e., mark by red) across different settings. Question 4 and 5 represernt the answer from LLaDA-MedV$^{V_1}$ and LLaDA-MedV$^{V_2}$, respectively. We omit the corresponding images for clarity. }
  \label{fig:reptition}

\end{figure}

The choice of initialization weights also appears to influence performance. Our experiments indicate that initializing LLaDA-MedV with LLaDA-V, which is trained on general image and video understanding tasks, results in suboptimal performance, even after supervised fine-tuning (Tab.~\ref{tab:compare-training}). Both LLaDA-MedV$^{V_1}$ (without alignment) and LLaDA-MedV$^{V_2}$ (with alignment) perform worse than LLaDA-MedV$^*$, suggesting that general-domain priors may not fully align with biomedical semantics. Additionally, as illustrated in question 4 and 5 of Fig.~\ref{fig:reptition}, these variants are more prone to generating repetitive responses under identical settings. Together, these findings highlight the importance of selecting domain-appropriate initialization strategies and incorporating fine-tuning to effectively adapt diffusion-based models to biomedical applications.

\begin{table}[ht]
    \centering
    \resizebox{\columnwidth}{!}{
    \begin{tabular}{c|cccc|c}
    \midrule
        \textbf{} & \multicolumn{4}{c|}{\textbf{Hyperparameters}} & \textbf{Open-end Conversation} \\
        \textbf{Group} & $L$ & $B$ & $Z$ & $Z\cdot B/L$ & Overall Score \\
    \midrule
        \multirow{4}{*}{(A)} & 32  & 32  & 32  & 32  & 37.057 \\
                             & 64  & 32  & 64  & 32  & 39.135 \\
                             & 128 & 32  & 128 & 32  & 33.222 \\
                             & 256 & 32  & 256 & 32  & 51.215 \\
    \midrule
        \multirow{4}{*}{(B)} & 256 & 256 & 32  & 32  & 20.165 \\
                             & 256 & 256 & 64  & 64  & 31.774 \\
                             & 256 & 256 & 128 & 128 & 44.020 \\
                             & 256 & 256 & 256 & 256 & 51.470 \\
    \midrule
        \multirow{4}{*}{(C)} & 256 & 32  & 256 & 32  & 51.215 \\
                             & \cellcolor[gray]{0.85}256 & \cellcolor[gray]{0.85}64  & \cellcolor[gray]{0.85}256 & \cellcolor[gray]{0.85}64  & \cellcolor[gray]{0.85}52.605 \\
                             & 256 & 128 & 256 & 128 & 51.641 \\
                             & 256 & 256 & 256 & 256 & 51.470 \\
    \midrule
    \end{tabular}}
    \caption{Performance under different hyperparameter configurations. Gray cells denote the setting outlined in Tab.~\ref{tab:chatbot}. $L$ represents the generation length, $B$ denotes the block length, and $Z$ indicates the total sampling steps. The term $Z \cdot B / L$ corresponds to the sampling steps per block.}
    \label{tab:compare-test}
\end{table}

\subsection{Inference-Time Key Factors}

We next examine how inference-time factors affect the quality and efficiency of generated responses. Specifically, we study the impact of generation length, the number of sampling steps, and block length within the semi-autoregressive generation process, where applicable. To investigate the effect of response length in real-world open-ended scenarios, we progressively increased the generation length $L$ while keeping the number of sampling steps per block fixed. As shown in Tab.~\ref{tab:compare-test}(A), we found out that increasing $L$ from 32 to 256 results in fluctuating performance, rather than the insensitiveness reported in previous work (e.g.,~\cite{nie2025large}). This discrepancy may initially criticize our experiments (i.e., Tab.~\ref{tab:chatbot}), as longer responses typically provide richer context to GPT-based evaluators, which could improve perceived answer quality. However, we emphasize that our evaluation is conducted under a controlled setting in which all models are encouraged to produce responses of fixed length (i.e., $L = 256$). For ARMs such as LLaVA-Med, this is enforced by setting the maximum token length to 256. In contrast, LLaDA-MedV's masked prediction mechanism inherently promotes the generation of more complete and informative responses under the same length constraint. Furthermore, differences in benchmark design and evaluation protocols may also contribute to the observed variation in performance trends.

Second, we observe that the number of sampling steps $Z$ introduces a trade-off between generation quality and computational efficiency. Generating a full response requires $Z$ sampling steps. Naturally, reducing $Z$ lowers the time cost associated with reconstructing the original content $r_0$. As shown in Tab.~\ref{tab:response-tw}, when $Z$ gradually decreases, the average generation time per word decreases 5 times with the compensation of clear performance drop (also noted in Tab.~\ref{tab:compare-test}(B)). This outcome is intuitive, as fewer steps introduce more noise into the reverse generative process described in Eq.~\ref{eq:backward-llada-main}.

Finally, we find that the block length $B$ requires careful tuning in the semi-autoregressive generation setting. When $B < L$, the model applies semi-autoregressive generation. However, as shown in Tab.~\ref{tab:compare-test}(C), increasing $B$, and thus increasing the number of sampling steps per block (i.e., $Z \cdot B / L$), does not consistently improve performance. This observation underscores the need for further investigation into optimal block configurations and their interaction with sampling dynamics. \textit{\textbf{Additional visualizations are provided in Appendix~B}}.

\subsection{Limitation and Future Works}

In addition to the findings discussed above, we observe a key limitation that merits further investigation: token repetition during generation. As shown in Fig.~\ref{fig:reptition}, the model exhibits a tendency to repeat tokens excessively under certain conditions. For instance, in Questions 2 and 3, the word “the” appears repeatedly until the target output length is reached. This issue becomes more pronounced when the desired response length is large, but the number of sampling steps is relatively small. One intuitive explanation is that when the number of sampling steps is insufficient, the model has fewer opportunities to remask and refine low-confidence tokens. Consequently, early generation errors or repetitive patterns are likely to persist throughout the generation process. Furthermore, because the response length $L$ imposes an upper bound on the number of total sampling steps (i.e., $Z \leq L$), this creates an inherent trade-off between generation quality and computational efficiency in long-form response settings.
These observations suggest a need for better strategies to balance $L$ and $Z$ during inference. Future work may explore adaptive step allocation or more effective remasking schedules to reduce repetition while preserving efficiency, especially in applications that require lengthy and detailed outputs.

\section{Conclusion}
In this work, we introduce LLaDA-MedV, the first large language diffusion model designed for biomedical image understanding. Leveraging visual instruction tuning, LLaDA-MedV demonstrates competitive performance relative to ARMs across both open-ended biomedical conversation tasks and downstream biomedical VQA, underscoring its potential as a reliable biomedical AI assistant. Our analysis further emphasizes the critical roles of of initialization weight selection, data-specific fine-tuning, and sampling step configuration in optimizing performance. Enhancing inference efficiency and mitigating token repetition represent important avenues for future research.

{
    \small
    \bibliographystyle{ieeenat_fullname}
    \bibliography{main}

@inproceedings{moor2023med,
  title={Med-flamingo: a multimodal medical few-shot learner},
  author={Moor, Michael and Huang, Qian and Wu, Shirley and Yasunaga, Michihiro and Dalmia, Yash and Leskovec, Jure and Zakka, Cyril and Reis, Eduardo Pontes and Rajpurkar, Pranav},
  booktitle={Machine Learning for Health (ML4H)},
  pages={353--367},
  year={2023},
  organization={PMLR}
}

@inproceedings{bannur2023learning,
  title={Learning to exploit temporal structure for biomedical vision-language processing},
  author={Bannur, Shruthi and Hyland, Stephanie and Liu, Qianchu and Perez-Garcia, Fernando and Ilse, Maximilian and Castro, Daniel C and Boecking, Benedikt and Sharma, Harshita and Bouzid, Kenza and Thieme, Anja and others},
  booktitle={Proceedings of the IEEE/CVF Conference on Computer Vision and Pattern Recognition},
  pages={15016--15027},
  year={2023}
}

@inproceedings{wang2022medclip,
  title={Medclip: Contrastive learning from unpaired medical images and text},
  author={Wang, Zifeng and Wu, Zhenbang and Agarwal, Dinesh and Sun, Jimeng},
  booktitle={Proceedings of the Conference on Empirical Methods in Natural Language Processing. Conference on Empirical Methods in Natural Language Processing},
  volume={2022},
  pages={3876},
  year={2022}
}

@inproceedings{lin2023pmc,
  title={Pmc-clip: Contrastive language-image pre-training using biomedical documents},
  author={Lin, Weixiong and Zhao, Ziheng and Zhang, Xiaoman and Wu, Chaoyi and Zhang, Ya and Wang, Yanfeng and Xie, Weidi},
  booktitle={International Conference on Medical Image Computing and Computer-Assisted Intervention},
  pages={525--536},
  year={2023},
  organization={Springer}
}

@article{lou2023discrete,
  title={Discrete diffusion modeling by estimating the ratios of the data distribution},
  author={Lou, Aaron and Meng, Chenlin and Ermon, Stefano},
  journal={arXiv preprint arXiv:2310.16834},
  year={2023}
}

@article{shi2024simplified,
  title={Simplified and generalized masked diffusion for discrete data},
  author={Shi, Jiaxin and Han, Kehang and Wang, Zhe and Doucet, Arnaud and Titsias, Michalis},
  journal={Advances in neural information processing systems},
  volume={37},
  pages={103131--103167},
  year={2024}
}

@article{sahoo2024simple,
  title={Simple and effective masked diffusion language models},
  author={Sahoo, Subham and Arriola, Marianne and Schiff, Yair and Gokaslan, Aaron and Marroquin, Edgar and Chiu, Justin and Rush, Alexander and Kuleshov, Volodymyr},
  journal={Advances in Neural Information Processing Systems},
  volume={37},
  pages={130136--130184},
  year={2024}
}

@article{ho2020denoising,
  title={Denoising diffusion probabilistic models},
  author={Ho, Jonathan and Jain, Ajay and Abbeel, Pieter},
  journal={Advances in neural information processing systems},
  volume={33},
  pages={6840--6851},
  year={2020}
}

@article{song2020denoising,
  title={Denoising diffusion implicit models},
  author={Song, Jiaming and Meng, Chenlin and Ermon, Stefano},
  journal={arXiv preprint arXiv:2010.02502},
  year={2020}
}

@article{song2020score,
  title={Score-based generative modeling through stochastic differential equations},
  author={Song, Yang and Sohl-Dickstein, Jascha and Kingma, Diederik P and Kumar, Abhishek and Ermon, Stefano and Poole, Ben},
  journal={arXiv preprint arXiv:2011.13456},
  year={2020}
}

@article{nie2025large,
  title={Large language diffusion models},
  author={Nie, Shen and Zhu, Fengqi and You, Zebin and Zhang, Xiaolu and Ou, Jingyang and Hu, Jun and Zhou, Jun and Lin, Yankai and Wen, Ji-Rong and Li, Chongxuan},
  journal={arXiv preprint arXiv:2502.09992},
  year={2025}
}

@article{grattafiori2024llama,
  title={The llama 3 herd of models},
  author={Grattafiori, Aaron and Dubey, Abhimanyu and Jauhri, Abhinav and Pandey, Abhinav and Kadian, Abhishek and Al-Dahle, Ahmad and Letman, Aiesha and Mathur, Akhil and Schelten, Alan and Vaughan, Alex and others},
  journal={arXiv preprint arXiv:2407.21783},
  year={2024}
}

@article{you2025llada,
  title={Llada-v: Large language diffusion models with visual instruction tuning},
  author={You, Zebin and Nie, Shen and Zhang, Xiaolu and Hu, Jun and Zhou, Jun and Lu, Zhiwu and Wen, Ji-Rong and Li, Chongxuan},
  journal={arXiv preprint arXiv:2505.16933},
  year={2025}
}

@article{liu2023visual,
  title={Visual instruction tuning},
  author={Liu, Haotian and Li, Chunyuan and Wu, Qingyang and Lee, Yong Jae},
  journal={Advances in neural information processing systems},
  volume={36},
  pages={34892--34916},
  year={2023}
}

@article{li2023llava,
  title={Llava-med: Training a large language-and-vision assistant for biomedicine in one day},
  author={Li, Chunyuan and Wong, Cliff and Zhang, Sheng and Usuyama, Naoto and Liu, Haotian and Yang, Jianwei and Naumann, Tristan and Poon, Hoifung and Gao, Jianfeng},
  journal={Advances in Neural Information Processing Systems},
  volume={36},
  pages={28541--28564},
  year={2023}
}

@article{lau2018dataset,
  title={A dataset of clinically generated visual questions and answers about radiology images},
  author={Lau, Jason J and Gayen, Soumya and Ben Abacha, Asma and Demner-Fushman, Dina},
  journal={Scientific data},
  volume={5},
  number={1},
  pages={1--10},
  year={2018},
  publisher={Nature Publishing Group}
}

@inproceedings{liu2021slake,
  title={Slake: A semantically-labeled knowledge-enhanced dataset for medical visual question answering},
  author={Liu, Bo and Zhan, Li-Ming and Xu, Li and Ma, Lin and Yang, Yan and Wu, Xiao-Ming},
  booktitle={2021 IEEE 18th international symposium on biomedical imaging (ISBI)},
  pages={1650--1654},
  year={2021},
  organization={IEEE}
}

@article{he2020pathvqa,
  title={Pathvqa: 30000+ questions for medical visual question answering},
  author={He, Xuehai and Zhang, Yichen and Mou, Luntian and Xing, Eric and Xie, Pengtao},
  journal={arXiv preprint arXiv:2003.10286},
  year={2020}
}

@article{li2022diffusion,
  title={Diffusion-lm improves controllable text generation},
  author={Li, Xiang and Thickstun, John and Gulrajani, Ishaan and Liang, Percy S and Hashimoto, Tatsunori B},
  journal={Advances in neural information processing systems},
  volume={35},
  pages={4328--4343},
  year={2022}
}

@article{gong2022diffuseq,
  title={Diffuseq: Sequence to sequence text generation with diffusion models},
  author={Gong, Shansan and Li, Mukai and Feng, Jiangtao and Wu, Zhiyong and Kong, LingPeng},
  journal={arXiv preprint arXiv:2210.08933},
  year={2022}
}

@article{strudel2022self,
  title={Self-conditioned embedding diffusion for text generation},
  author={Strudel, Robin and Tallec, Corentin and Altch{\'e}, Florent and Du, Yilun and Ganin, Yaroslav and Mensch, Arthur and Grathwohl, Will and Savinov, Nikolay and Dieleman, Sander and Sifre, Laurent and others},
  journal={arXiv preprint arXiv:2211.04236},
  year={2022}
}

@article{chen2022analog,
  title={Analog bits: Generating discrete data using diffusion models with self-conditioning},
  author={Chen, Ting and Zhang, Ruixiang and Hinton, Geoffrey},
  journal={arXiv preprint arXiv:2208.04202},
  year={2022}
}

@article{richemond2022categorical,
  title={Categorical sdes with simplex diffusion},
  author={Richemond, Pierre H and Dieleman, Sander and Doucet, Arnaud},
  journal={arXiv preprint arXiv:2210.14784},
  year={2022}
}

@article{ye2023dinoiser,
  title={Dinoiser: Diffused conditional sequence learning by manipulating noises},
  author={Ye, Jiasheng and Zheng, Zaixiang and Bao, Yu and Qian, Lihua and Wang, Mingxuan},
  journal={arXiv preprint arXiv:2302.10025},
  year={2023}
}

@article{mahabadi2023tess,
  title={Tess: Text-to-text self-conditioned simplex diffusion},
  author={Mahabadi, Rabeeh Karimi and Ivison, Hamish and Tae, Jaesung and Henderson, James and Beltagy, Iz and Peters, Matthew E and Cohan, Arman},
  journal={arXiv preprint arXiv:2305.08379},
  year={2023}
}

@article{ou2024your,
  title={Your absorbing discrete diffusion secretly models the conditional distributions of clean data},
  author={Ou, Jingyang and Nie, Shen and Xue, Kaiwen and Zhu, Fengqi and Sun, Jiacheng and Li, Zhenguo and Li, Chongxuan},
  journal={arXiv preprint arXiv:2406.03736},
  year={2024}
}

@article{gong2024scaling,
  title={Scaling diffusion language models via adaptation from autoregressive models},
  author={Gong, Shansan and Agarwal, Shivam and Zhang, Yizhe and Ye, Jiacheng and Zheng, Lin and Li, Mukai and An, Chenxin and Zhao, Peilin and Bi, Wei and Han, Jiawei and others},
  journal={arXiv preprint arXiv:2410.17891},
  year={2024}
}

@article{zhu2025retinalgpt,
  title={RetinalGPT: A Retinal Clinical Preference Conversational Assistant Powered by Large Vision-Language Models},
  author={Zhu, Wenhui and Li, Xin and Chen, Xiwen and Qiu, Peijie and Krishna Vasa, Vamsi and Dong, Xuanzhao and Chen, Yanxi and Lepore, Natasha and Dumitrascu, Oana and Su, Yi and others},
  journal={arXiv e-prints},
  pages={arXiv--2503},
  year={2025}
}

@article{zhu2025toward,
  title={Toward Effective Reinforcement Learning Fine-Tuning for Medical VQA in Vision-Language Models},
  author={Zhu, Wenhui and Dong, Xuanzhao and Li, Xin and Qiu, Peijie and Chen, Xiwen and Razi, Abolfazl and Sotiras, Aris and Su, Yi and Wang, Yalin},
  journal={arXiv e-prints},
  pages={arXiv--2505},
  year={2025}
}

@article{luo2023biomedgpt,
  title={Biomedgpt: Open multimodal generative pre-trained transformer for biomedicine},
  author={Luo, Yizhen and Zhang, Jiahuan and Fan, Siqi and Yang, Kai and Wu, Yushuai and Qiao, Mu and Nie, Zaiqing},
  journal={arXiv preprint arXiv:2308.09442},
  year={2023}
}

@article{zhang2023biomedgpt,
  title={Biomedgpt: A unified and generalist biomedical generative pre-trained transformer for vision, language, and multimodal tasks},
  author={Zhang, Kai and Yu, Jun and Adhikarla, Eashan and Zhou, Rong and Yan, Zhiling and Liu, Yixin and Liu, Zhengliang and He, Lifang and Davison, Brian and Li, Xiang and others},
  journal={arXiv e-prints},
  pages={arXiv--2305},
  year={2023}
}

@article{pan2025medvlm,
  title={Medvlm-r1: Incentivizing medical reasoning capability of vision-language models (vlms) via reinforcement learning},
  author={Pan, Jiazhen and Liu, Che and Wu, Junde and Liu, Fenglin and Zhu, Jiayuan and Li, Hongwei Bran and Chen, Chen and Ouyang, Cheng and Rueckert, Daniel},
  journal={arXiv preprint arXiv:2502.19634},
  year={2025}
}

@article{lai2025med,
  title={Med-r1: Reinforcement learning for generalizable medical reasoning in vision-language models},
  author={Lai, Yuxiang and Zhong, Jike and Li, Ming and Zhao, Shitian and Yang, Xiaofeng},
  journal={arXiv preprint arXiv:2503.13939},
  year={2025}
}

@article{denner2024visual,
  title={Visual Prompt Engineering for Vision Language Models in Radiology},
  author={Denner, Stefan and Bujotzek, Markus and Bounias, Dimitrios and Zimmerer, David and Stock, Raphael and Maier-Hein, Klaus},
  journal={arXiv preprint arXiv:2408.15802},
  year={2024}
}

@article{wang2024does,
  title={How Does Diverse Interpretability of Textual Prompts Impact Medical Vision-Language Zero-Shot Tasks?},
  author={Wang, Sicheng and Liu, Che and Arcucci, Rossella},
  journal={arXiv preprint arXiv:2409.00543},
  year={2024}
}

@article{guo2024prompting,
  title={Prompting medical large vision-language models to diagnose pathologies by visual question answering},
  author={Guo, Danfeng and Terzopoulos, Demetri},
  journal={arXiv preprint arXiv:2407.21368},
  year={2024}
}

@article{wang2024interactive,
  title={Interactive computer-aided diagnosis on medical image using large language models},
  author={Wang, Sheng and Zhao, Zihao and Ouyang, Xi and Liu, Tianming and Wang, Qian and Shen, Dinggang},
  journal={Communications Engineering},
  volume={3},
  number={1},
  pages={133},
  year={2024},
  publisher={Nature Publishing Group UK London}
}

@article{austin2021structured,
  title={Structured denoising diffusion models in discrete state-spaces},
  author={Austin, Jacob and Johnson, Daniel D and Ho, Jonathan and Tarlow, Daniel and Van Den Berg, Rianne},
  journal={Advances in neural information processing systems},
  volume={34},
  pages={17981--17993},
  year={2021}
}

@article{vincent2011connection,
  title={A connection between score matching and denoising autoencoders},
  author={Vincent, Pascal},
  journal={Neural computation},
  volume={23},
  number={7},
  pages={1661--1674},
  year={2011},
  publisher={MIT Press}
}

@article{touvron2023llama,
  title={Llama: Open and efficient foundation language models},
  author={Touvron, Hugo and Lavril, Thibaut and Izacard, Gautier and Martinet, Xavier and Lachaux, Marie-Anne and Lacroix, Timoth{\'e}e and Rozi{\`e}re, Baptiste and Goyal, Naman and Hambro, Eric and Azhar, Faisal and others},
  journal={arXiv preprint arXiv:2302.13971},
  year={2023}
}

@inproceedings{radford2021learning,
  title={Learning transferable visual models from natural language supervision},
  author={Radford, Alec and Kim, Jong Wook and Hallacy, Chris and Ramesh, Aditya and Goh, Gabriel and Agarwal, Sandhini and Sastry, Girish and Askell, Amanda and Mishkin, Pamela and Clark, Jack and others},
  booktitle={International conference on machine learning},
  pages={8748--8763},
  year={2021},
  organization={PmLR}
}

@inproceedings{liu2024improved,
  title={Improved baselines with visual instruction tuning},
  author={Liu, Haotian and Li, Chunyuan and Li, Yuheng and Lee, Yong Jae},
  booktitle={Proceedings of the IEEE/CVF conference on computer vision and pattern recognition},
  pages={26296--26306},
  year={2024}
}

@article{tschannen2025siglip,
  title={Siglip 2: Multilingual vision-language encoders with improved semantic understanding, localization, and dense features},
  author={Tschannen, Michael and Gritsenko, Alexey and Wang, Xiao and Naeem, Muhammad Ferjad and Alabdulmohsin, Ibrahim and Parthasarathy, Nikhil and Evans, Talfan and Beyer, Lucas and Xia, Ye and Mustafa, Basil and others},
  journal={arXiv preprint arXiv:2502.14786},
  year={2025}
}

@inproceedings{chang2022maskgit,
  title={Maskgit: Masked generative image transformer},
  author={Chang, Huiwen and Zhang, Han and Jiang, Lu and Liu, Ce and Freeman, William T},
  booktitle={Proceedings of the IEEE/CVF conference on computer vision and pattern recognition},
  pages={11315--11325},
  year={2022}
}

@misc{openai2023gpt4,
  title        = {GPT-4 Technical Report},
  author       = {OpenAI and Achiam, Josh and Adler, Steven and Agarwal, Sandhini and Ahmad, Lama and Akkaya, Ilge and … and Zoph, Barret},
  year         = {2023},
  howpublished = {\url{https://arxiv.org/abs/2303.08774}},
  note         = {arXiv:2303.08774, submitted March 15 2023} 
}

@misc{openai2025gpt41mini,
  title        = {Introducing GPT-4.1 in the API},
  author       = {OpenAI},
  year         = {2025},
  month        = apr,
  day          = {14},
  howpublished = {\url{https://openai.com/index/gpt-4-1/}},
  note         = {Includes GPT‑4.1, GPT‑4.1 mini, and GPT‑4.1 nano}
}

@article{guo2025deepseek,
  title={Deepseek-r1: Incentivizing reasoning capability in llms via reinforcement learning},
  author={Guo, Daya and Yang, Dejian and Zhang, Haowei and Song, Junxiao and Zhang, Ruoyu and Xu, Runxin and Zhu, Qihao and Ma, Shirong and Wang, Peiyi and Bi, Xiao and others},
  journal={arXiv preprint arXiv:2501.12948},
  year={2025}
}

@article{ouyang2022training,
  title={Training language models to follow instructions with human feedback},
  author={Ouyang, Long and Wu, Jeffrey and Jiang, Xu and Almeida, Diogo and Wainwright, Carroll and Mishkin, Pamela and Zhang, Chong and Agarwal, Sandhini and Slama, Katarina and Ray, Alex and others},
  journal={Advances in neural information processing systems},
  volume={35},
  pages={27730--27744},
  year={2022}
}

@article{hsu2023gpt,
  title={Gpt-4 as an effective zero-shot evaluator for scientific figure captions},
  author={Hsu, Ting-Yao and Huang, Chieh-Yang and Rossi, Ryan and Kim, Sungchul and Giles, C Lee and Huang, Ting-Hao K},
  journal={arXiv preprint arXiv:2310.15405},
  year={2023}
}

@article{yang2024gpt,
  title={Gpt-4 as evaluator: Evaluating large language models on pest management in agriculture},
  author={Yang, Shanglong and Yuan, Zhipeng and Li, Shunbao and Peng, Ruoling and Liu, Kang and Yang, Po},
  journal={arXiv preprint arXiv:2403.11858},
  year={2024}
}

@article{liu2023g,
  title={G-eval: NLG evaluation using gpt-4 with better human alignment},
  author={Liu, Yang and Iter, Dan and Xu, Yichong and Wang, Shuohang and Xu, Ruochen and Zhu, Chenguang},
  journal={arXiv preprint arXiv:2303.16634},
  year={2023}
}

@article{Qwen-VL,
  title={Qwen-VL: A Versatile Vision-Language Model for Understanding, Localization, Text Reading, and Beyond},
  author={Bai, Jinze and Bai, Shuai and Yang, Shusheng and Wang, Shijie and Tan, Sinan and Wang, Peng and Lin, Junyang and Zhou, Chang and Zhou, Jingren},
  journal={arXiv preprint arXiv:2308.12966},
  year={2023}
}

@misc{meta2024llama3.2-11b-vision,
  title        = {{Llama-3.2-11B-Vision: A Multimodal Vision--Language LLM}},
  author       = {{Meta AI}},
  year         = {2024},
  month        = sep,
  day          = {25},
  howpublished = {Model card via Meta AI},
  note         = {Version released September 25, 2024; instruction-tuned for image reasoning, captioning, and VQA with 10.6B parameters}
}

@article{bazi2023vision,
  title={Vision--language model for visual question answering in medical imagery},
  author={Bazi, Yakoub and Rahhal, Mohamad Mahmoud Al and Bashmal, Laila and Zuair, Mansour},
  journal={Bioengineering},
  volume={10},
  number={3},
  pages={380},
  year={2023},
  publisher={MDPI}
}

@inproceedings{liu2023q2atransformer,
  title={Q2atransformer: Improving medical vqa via an answer querying decoder},
  author={Liu, Yunyi and Wang, Zhanyu and Xu, Dong and Zhou, Luping},
  booktitle={International conference on information processing in medical imaging},
  pages={445--456},
  year={2023},
  organization={Springer}
}

@inproceedings{van2023open,
  title={Open-ended medical visual question answering through prefix tuning of language models},
  author={Van Sonsbeek, Tom and Derakhshani, Mohammad Mahdi and Najdenkoska, Ivona and Snoek, Cees GM and Worring, Marcel},
  booktitle={International Conference on Medical Image Computing and Computer-Assisted Intervention},
  pages={726--736},
  year={2023},
  organization={Springer}
}

@inproceedings{eslami2023pubmedclip,
  title={Pubmedclip: How much does clip benefit visual question answering in the medical domain?},
  author={Eslami, Sedigheh and Meinel, Christoph and De Melo, Gerard},
  booktitle={Findings of the Association for Computational Linguistics: EACL 2023},
  pages={1181--1193},
  year={2023}
}

@article{zhang2023large,
  title={Large-scale domain-specific pretraining for biomedical vision-language processing},
  author={Zhang, Sheng and Xu, Yanbo and Usuyama, Naoto and Bagga, Jaspreet and Tinn, Robert and Preston, Sam and Rao, Rajesh and Wei, Mu and Valluri, Naveen and Wong, Cliff and others},
  journal={arXiv preprint arXiv:2303.00915},
  volume={2},
  number={3},
  pages={6},
  year={2023}
}

@inproceedings{li2023self,
  title={Self-supervised vision-language pretraining for medial visual question answering},
  author={Li, Pengfei and Liu, Gang and Tan, Lin and Liao, Jinying and Zhong, Shenjun},
  booktitle={2023 IEEE 20th International Symposium on Biomedical Imaging (ISBI)},
  pages={1--5},
  year={2023},
  organization={IEEE}
}

@inproceedings{rajbhandari2021zero,
  title={Zero-infinity: Breaking the gpu memory wall for extreme scale deep learning},
  author={Rajbhandari, Samyam and Ruwase, Olatunji and Rasley, Jeff and Smith, Shaden and He, Yuxiong},
  booktitle={Proceedings of the international conference for high performance computing, networking, storage and analysis},
  pages={1--14},
  year={2021}
}
}


\clearpage
\appendix
\section{Implementation Details}
\subsection{Training Configuration}
The training process of LLaDA-MedV is structured into three stages. Following the approach of~\cite{li2023llava}, the first two stages aim to establish semantic alignment between biomedical language and visual content, while also enabling the model to follow visual instructions within a biomedical context. To further improve performance in dataset-specific scenarios, we introduce a third stage involving supervised fine-tuning (SFT) on three biomedical VQA datasets. 

Throughout all stages, we employ LLaDA-8B-Instruct~\cite{nie2025large} as the language backbone. The vision tower is based on SigLIP2~\cite{tschannen2025siglip}, and the vision-language projection module is implemented as a lightweight two-layer MLP with GELU activation.  All training is conducted using four NVIDIA A100 GPUs (80GB each). Additional training details are provided in Tab.~\ref{tab:training_config}.
\begin{table*}[th]
\centering
\resizebox{1\textwidth}{!}{\begin{tabular}{l|c|c|c|c|c}
\midrule
\textbf{Training stage} & Alignment & MD-SFT  & SD-SFT (VQA-RAD)  & SD-SFT(SLAKE) & SD-SFT(PathVQA) \\
\midrule
Vision tower & \multicolumn{5}{c}{Siglip2-so400m-patch14-384~\cite{tschannen2025siglip}} \\
Language tower & \multicolumn{5}{c}{LLaDA-8B-Instruct~\cite{nie2025large}} \\
Projector & \multicolumn{5}{c}{2-layer MLP with GELU}\\
Attention & \multicolumn{5}{c}{Bidirectional attention} \\
DeepSpeed & \multicolumn{5}{c}{ZeRO-3~\cite{rajbhandari2021zero}}\\
Optimizer & \multicolumn{5}{c}{AdamW} \\
Scheduler & \multicolumn{5}{c}{Cosine scheduler with 3\% warmup} \\
\midrule
Batch size (Global) & 32 & 8 & 8 & 8 & 8  \\
Model max length & 8192 & 8192 & 8192 & 8192 & 8192 \\
\#Samples in training set & 600K & 60k & 1797 & 4919 & 19755 \\
\midrule
LR of language tower & - & $1 \times 10^{-5}$ & $2 \times 10^{-6}$ & $2 \times 10^{-6}$ & $2 \times 10^{-6}$ \\
LR of projector & $1 \times 10^{-3}$ & $1 \times 10^{-5}$ & $2 \times 10^{-6}$ & $2 \times 10^{-6}$ & $2 \times 10^{-6}$\\
Epoch & 2 & 4 & 2 & 10 & 7 \\
\midrule
\end{tabular}}
\caption{Training configurations of LLaDA-MedV across three stages. \textbf{MD-SFT} denotes the multi-turn dialogue SFT, and \textbf{SD-SFT} represent the single-turn dialogue over training set of VQA-RAD, SLAKE and PathVQA, respectively.}
\label{tab:training_config}
\end{table*}

\subsection{Inference Procedure}
\paragraph{Inference Details.} The response generation process simulates the reverse dynamics of mask diffusion. Starting from a fully masked response (i.e., $t=1$), the learned mask predictor $p_\theta$ iteratively reconstructs the assistant's response. To better align with the reverse diffusion process, an appropriate remasking strategy is applied at each step. The detailed inference procedure is presented in Algorithm~\ref{alg:llada-basic}.
\begin{algorithm}[h]
\caption{Inference Strategy in LLaDA-MedV}
\label{alg:llada-basic}
\begin{algorithmic}[1]
  \REQUIRE Trained mask predictor $p_\theta$, user visual input $X_v$, text prompt $u_0$, response length $L$, and total sampling steps $Z$  
  \ENSURE Start with fully masked response $r_1$ of length $L$ (i.e., $t=1$)
  \FOR{$t$ from $1$ down to $1/Z$ with step $1/Z$}
    \STATE Set $s = t - 1/Z$
    \STATE Predict $r_0 = \arg\max p_\theta(r_0 \mid X_v, u_0, r_t)$
    \FOR{$i = 1$ to $L$}
      \IF{$r_t^i \ne \mathbf{M}$}
        \STATE Set $r_0^i = r_t^i$  \hfill \COMMENT{Retain meaningful token}
      \ELSE
        \STATE Remask $r_0^i = \mathbf{M}$ with probability $s/t$
      \ENDIF
    \ENDFOR
  \ENDFOR
  \RETURN $r_0$
\end{algorithmic}
\end{algorithm}

Following prior works~\cite{you2025llada,nie2025large}, we adopt the low-confidence remasking strategy~\cite{chang2022maskgit}, in which only tokens with low confidence (i.e., lower $p_\theta$ values) are remasked at each step (e.g., Step 8), rather than selecting tokens uniformly at random. In addition, we explore a semi-autoregressive generation strategy~\cite{nie2025large}. Specifically, when $B < L$, the response of length $L$ is divided into $L/B$ blocks, where $B$ denotes the block length. Generation proceeds sequentially from left to right across blocks, with each block undergoing $Z \cdot B / L$ sampling steps (i.e., as noted in Algorithm.~\ref{alg:llada-basic}). 

For the main open-ended biomedical conversation tasks, we apply this strategy with $L = 256$, $B = 64$, and $Z = 256$. However, for downstream biomedical VQA tasks, we disable the semi-autoregressive mechanism by setting $L = B = Z = 64$. Unless otherwise specified, this setting is used consistently in all detailed analyses.

\paragraph{Baseline Model Configuration.} We compare LLaDA-MedV against nine baseline vision-language models, covering both general-domain and biomedical-specific systems. These models are evaluated using their default system prompts and a maximum token limit of 256. For fairness, models with intermediate reasoning capabilities (e.g., MedVLM-R1) are evaluated using the full generated response.

In the open-ended biomedical conversation setting, we use GPT-4.1 mini~\cite{openai2025gpt41mini} to assess performance, as the original GPT-4 (0314) used in~\cite{li2023llava} is no longer publicly accessible. Additionally, Tab.~\ref{tab:baseline-summary} provides an overview of all baseline models included in this evaluation. For downstream VQA benchmarks, we report baseline results directly from~\cite{li2023llava} when available, and therefore omit redundant model-specific details for clarity. We kindly refer readers to~\cite{li2023llava} for comprehensive baseline configurations in the VQA setting. 
\begin{table*}[ht]
\centering
\caption{Overview of baseline models evaluated on the open-ended biomedical conversation task. The \textbf{Domain} column indicates whether each model is general-purpose or biomedical-specific. The \textbf{Size} column refers to the model size, where specified. Detailed model variants are included when available.}
\label{tab:baseline-summary}
\resizebox{0.8\linewidth}{!}{
\begin{tabular}{l|c|c|c}
\midrule
\textbf{Model} & \textbf{Domain} & \textbf{Size} & \textbf{Variant / Backbone} \\
\midrule
LLaMA~\cite{meta2024llama3.2-11b-vision} & General & 11B & LLaMA-3.2-11B-Vision  \\
LLaVA~\cite{liu2023visual} & General & 7B & llava-1.5-7B  \\
LLaVA-Med~\cite{li2023llava} & Biomedical & 7B & llava-med-v1.5-mistral-7B  \\
Med-Flamingo~\cite{moor2023med} & Biomedical & 9B & Med-Flamingo-9B \\
MedVLM-R1~\cite{pan2025medvlm} & Biomedical & - & -  \\
Qwen-RL-Med~\cite{zhu2025toward} & Biomedical & - & - \\
Qwen-VL~\cite{Qwen-VL} & General & 2B & Qwen2-VL-2B-Instruct  \\
RetinalGPT~\cite{zhu2025retinalgpt} & Biomedical & - & - \\
LLaDA-V~\cite{you2025llada} & General & - & -  \\
\midrule
\end{tabular}
}
\end{table*}

\subsection{Dataset}
\paragraph{Training dataset.} We adopt the training data introduced in~\cite{li2023llava} to equip LLaDA-MedV with biomedical visual understanding capabilities across three training stages. All images are processed using the SigLIP2 vision encoder~\cite{tschannen2025siglip}.
\begin{itemize}
    \item \textbf{Stage 1.} We use 600K aligned image-text pairs to train the vision-language projection module and learn robust cross-modal representations. Each training instance is structured as a single-turn dialogue, where the human prompt is removed except for a special \texttt{<image>} token. This setup ensures that the model learns to generate responses conditioned solely on image embeddings, without reliance on textual input.
    \item \textbf{Stage 2.} We leverage a 60K multi-turn dialogue dataset with inline entity mentions. Each instance contains a full conversation between a human and the assistant, where image embeddings are prepended only to the first human input. This configuration enables the model to follow visually grounded instructions across multiple dialogue turns.
    \item \textbf{Stage 3.} We perform supervised fine-tuning using the training sets from three biomedical VQA datasets: VQA-RAD~\cite{lau2018dataset}, SLAKE~\cite{liu2021slake}, and PathVQA~\cite{he2020pathvqa}. Each VQA sample is converted into a one-turn dialogue, where the question is treated as the human input and the ground-truth answer is used as the assistant's response. This training simulates the multi-turn dialogue structure introduced in Stage 2.
\end{itemize}

\paragraph{Open-end Biomedical Conversation Benchmark.}
We use the open-ended biomedical conversation benchmark proposed in~\cite{li2023llava} to evaluate model performance in realistic scenarios. Specifically, the dataset contains 193 novel questions paired with 50 corresponding images. Notably, these 50 image-caption pairs are entirely unseen during training. The questions fall into two categories: open-ended conversations and detailed descriptions. Conversation-style questions are typically broad biomedical queries related to the image content, while detailed description questions explicitly ask for fine-grained analysis of the visual information. As previously mentioned, ground-truth responses are generated by GPT-4 (i.e., GPT-4-0314) based solely on the associated figure captions. To evaluate the candidate model's response (e.g., from LLaDA-MedV), we use GPT-4.1 mini as the automatic evaluator. GPT-4.1 mini is provided with both the candidate response and the GPT-4 reference response, along with the original question, image caption, and relevant context. It is then prompted to rate each response across four dimensions, including helpfulness, relevance, accuracy, and level of detail, and to assign an overall score on a scale from 1 to 10, where higher scores indicate better performance. Additionally, GPT-4.1 mini provides a detailed rationale for each evaluation, facilitating more transparent comparisons. The final score for the candidate model is normalized by the GPT-4 reference score to compute a relative performance metric.

\paragraph{Downstream Biomedical Visual Question Answering.}
Following~\cite{li2023llava}, we use three biomedical visual question answering (VQA) benchmarks to further improve LLaDA-MedV performance in scenario that require high service quality. We outlined the detailed statistics in Tab.~\ref{tab:statistics}.
\begin{itemize}
    \item \textbf{VQA-RAD~\cite{lau2018dataset}} consists of 3,515 clinician-generated QA pairs grounded in 315 radiology images, which are evenly distributed across three anatomical regions: head, chest, and abdomen. Each image is paired with multiple questions, covering a diverse range of clinical inquiries. The questions are categorized into 11 distinct types, including abnormality, attribute, modality, organ system, color, counting, object or condition presence, size, imaging plane, positional reasoning, and others. Approximately half of the answers are closed-ended (i.e., yes/no), while the remaining are open-ended, typically expressed as a single word or short phrase. It is worth noting that we use the cleaned version of the dataset in all experiments, resulting in a total of 2,248 QA pairs.
    \item \textbf{SLAKE~\cite{liu2021slake}} is a Semantically-Labeled Knowledge-Enhanced benchmark designed for medical visual question answering. It contains 642 radiology images and over 7,000 diverse question-answer pairs, all annotated by experienced physicians. Many of the questions require external medical knowledge, which is supported by an accompanying medical knowledge graph. The dataset also provides rich visual annotations, including semantic segmentation masks and object detection bounding boxes. In addition, SLAKE encompasses a broader range of imaging modalities and anatomical regions compared to existing datasets, covering areas such as the brain, neck, chest, abdomen, and pelvic cavity. Here, we only consider the English subset throughout the experiments.
    \item \textbf{PathVQA~\cite{he2020pathvqa}} is a VQA benchmark consisting of pathology images. It includes 4,998 images accompanied by 32,799 question-answer pairs. Each image is associated with multiple questions addressing various aspects such as location, shape, color, and appearance. The questions are categorized into two main types: open-form (e.g., "what," "why," "how," "where") and closed-form, encompassing a wide range of question styles relevant to pathology interpretation.
\end{itemize}
We evaluate the performance of LLaDA-MedV and report baseline results outlined in~\cite{li2023llava} across all three VQA benchmarks, where available. However, for more detailed analysis, we focus exclusively on VQA-RAD and SLAKE to ensure evaluation efficiency.
\begin{table}[ht]
    \centering
    \resizebox{\linewidth}{!}{
    \begin{tabular}{l|cc|ccc|ccc}
    \midrule
     & \multicolumn{2}{c|}{\textbf{VQA-RAD}} & \multicolumn{3}{c|}{\textbf{SLAKE}} & \multicolumn{3}{c}{\textbf{PathVQA}} \\
     \textbf{Dataset Details}& Train & Test & Train & Val & Test & Train & Val & Test \\
     \midrule
      \# Images & 313 & 203 & 450 & 96 & 96 & 2599 & 858 & 858  \\
    \midrule
      \# QA & 1797 & 451 & 4919 & 1053 & 1061 & 19755 & 6279 & 6761 \\ 
      \# Closed-form  & 1027 & 272 & 1943 & 422 & 416 & 9806 & 3135 & 3391 \\
      \# Open-form & 770 & 179 & 2976 & 631 & 645 & 9949 & 3144 & 3370 \\
    \midrule
    \end{tabular}}
    \caption{Statistics of the downstream VQA benchmarks. \textbf{\#} denotes the number of samples in each split.}
    \label{tab:statistics}
\end{table}

\section{ Visualization Analysis}
\subsection{Open-end Biomedical Conversation}
\begin{figure}[ht]
  \centering
  \includegraphics[width=1\linewidth]{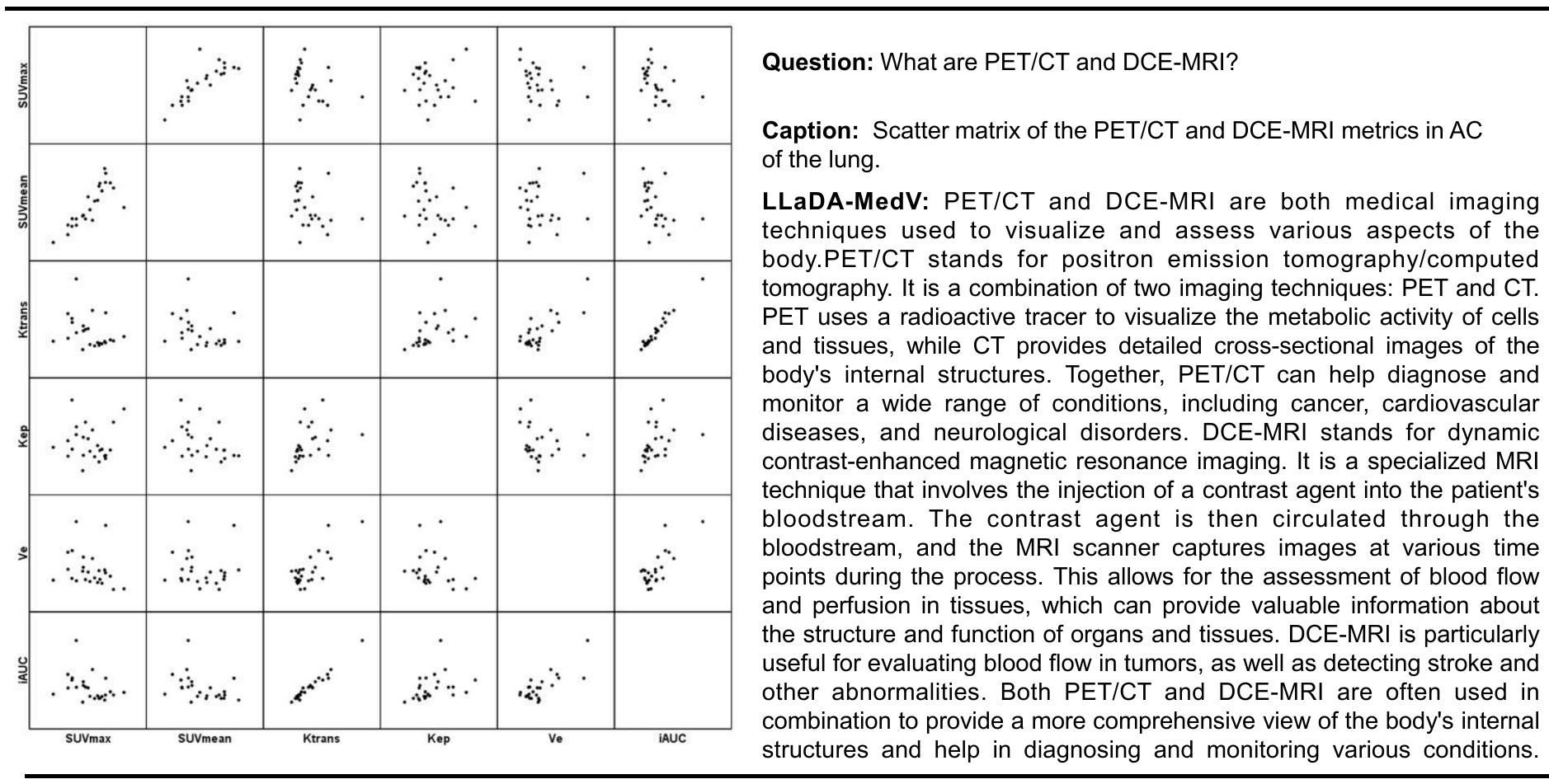}
  \caption{Illustration of LLaDA-MedV response in open-end biomedical conversation task. Image, question and corresponding caption comes from ~\cite{li2023llava}}
  \label{fig:oe-1}
\end{figure}

\begin{figure}[ht]
  \centering
  \includegraphics[width=1\linewidth]{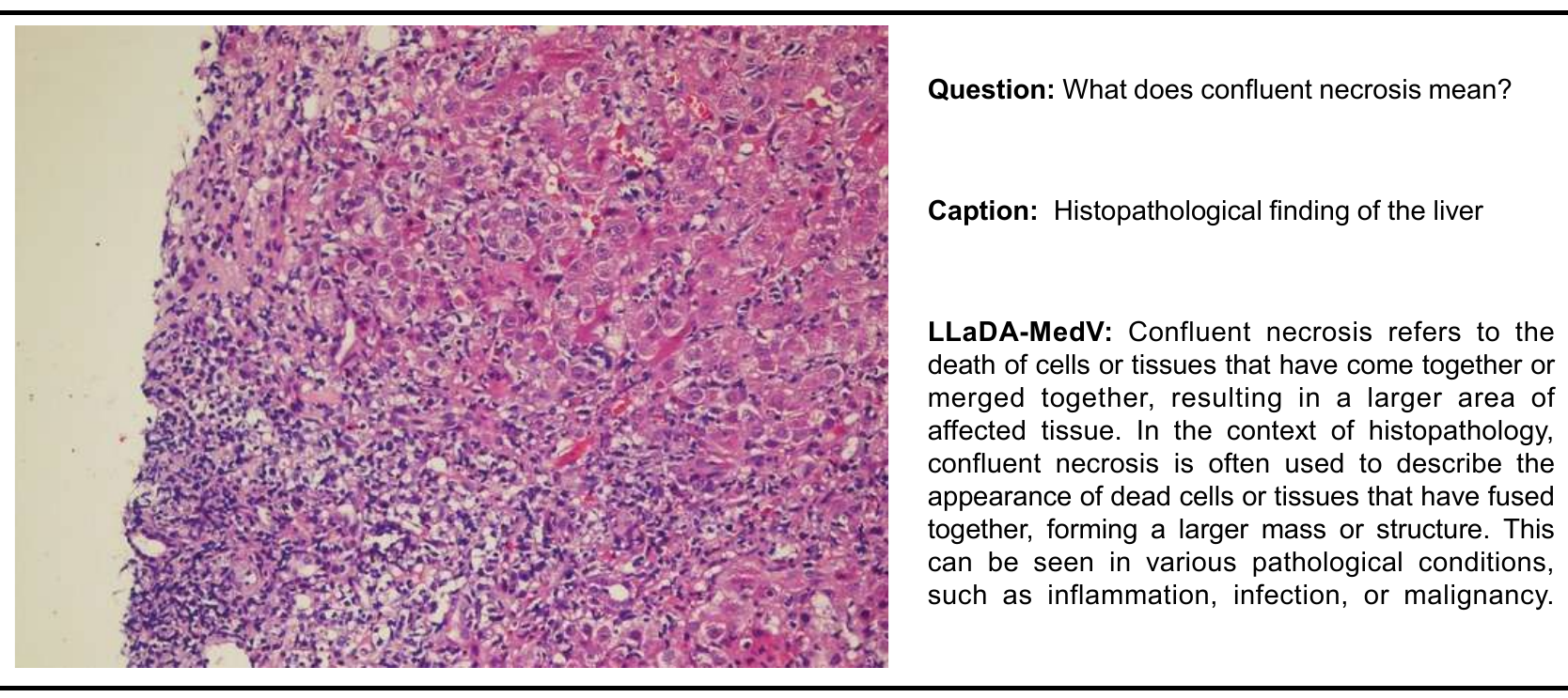}
  \caption{Illustration of LLaDA-MedV response in open-end biomedical conversation task. Image, question and corresponding caption comes from ~\cite{li2023llava}}
  \label{fig:oe-2}
\end{figure}

\begin{figure}[ht]
  \centering
  \includegraphics[width=1\linewidth]{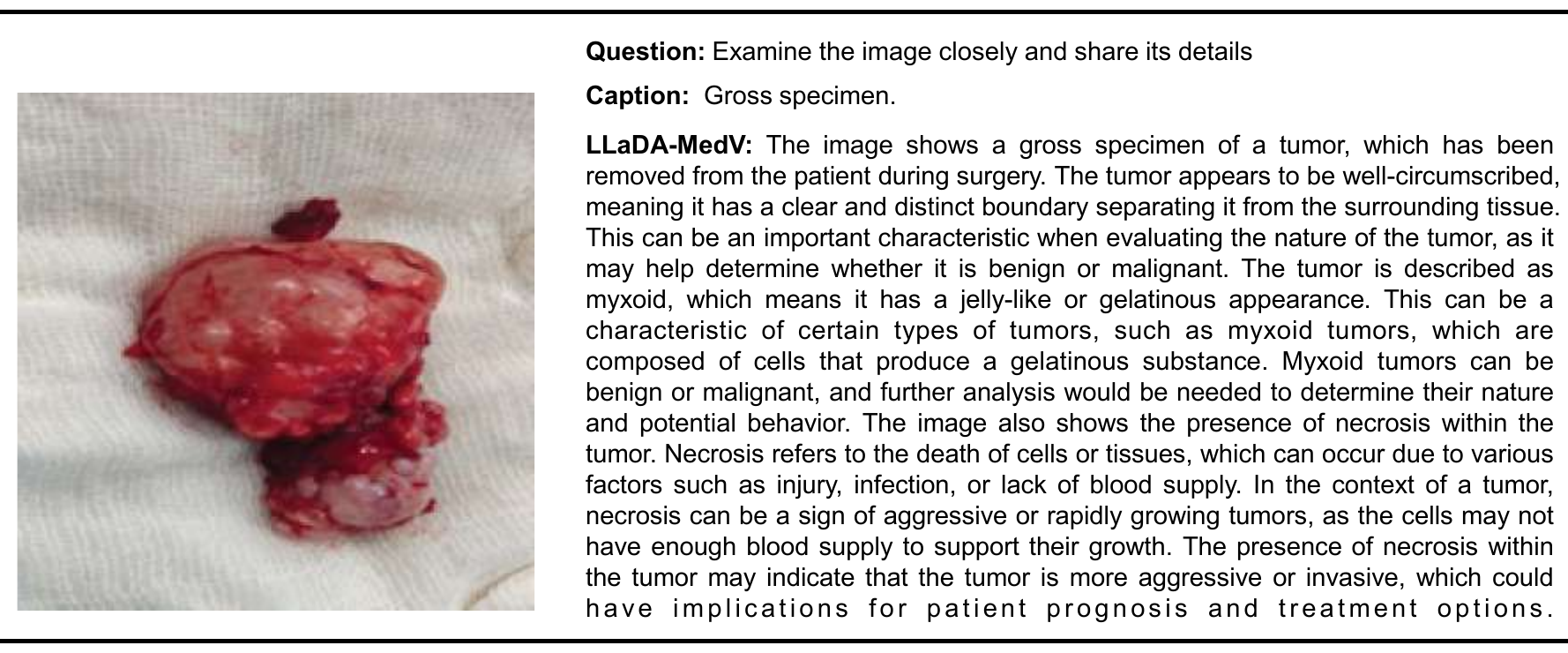}
  \caption{Illustration of LLaDA-MedV response in open-end biomedical conversation task. Image, question and corresponding caption comes from ~\cite{li2023llava}}
  \label{fig:oe-3}
\end{figure}

Due to space limitations in the main text, we provide additional visualizations for open-ended biomedical conversations in Fig.~\ref{fig:oe-1}, Fig.~\ref{fig:oe-2}, and Fig.~\ref{fig:oe-3}. As shown, LLaDA-MedV consistently produces relatively longer responses enriched with more contextual information. For all visualized examples, we use the following generation settings (as used in main setting): $L = 256$, $B = 64$, and $Z = 256$. For clarity, we omit GPT-4 reference answer.

\subsection{Inference-Time Key Factors}
In our investigation of inference-time factors influencing LLaDA-MedV, we systematically evaluate the effects of response length ($L$), block length ($B$), and the number of sampling steps ($Z$) on final generation quality. This section is dedicated to providing additional visualizations to facilitate a deeper understanding of their impact.
\begin{figure}[ht]
  \centering
  \includegraphics[width=1\linewidth]{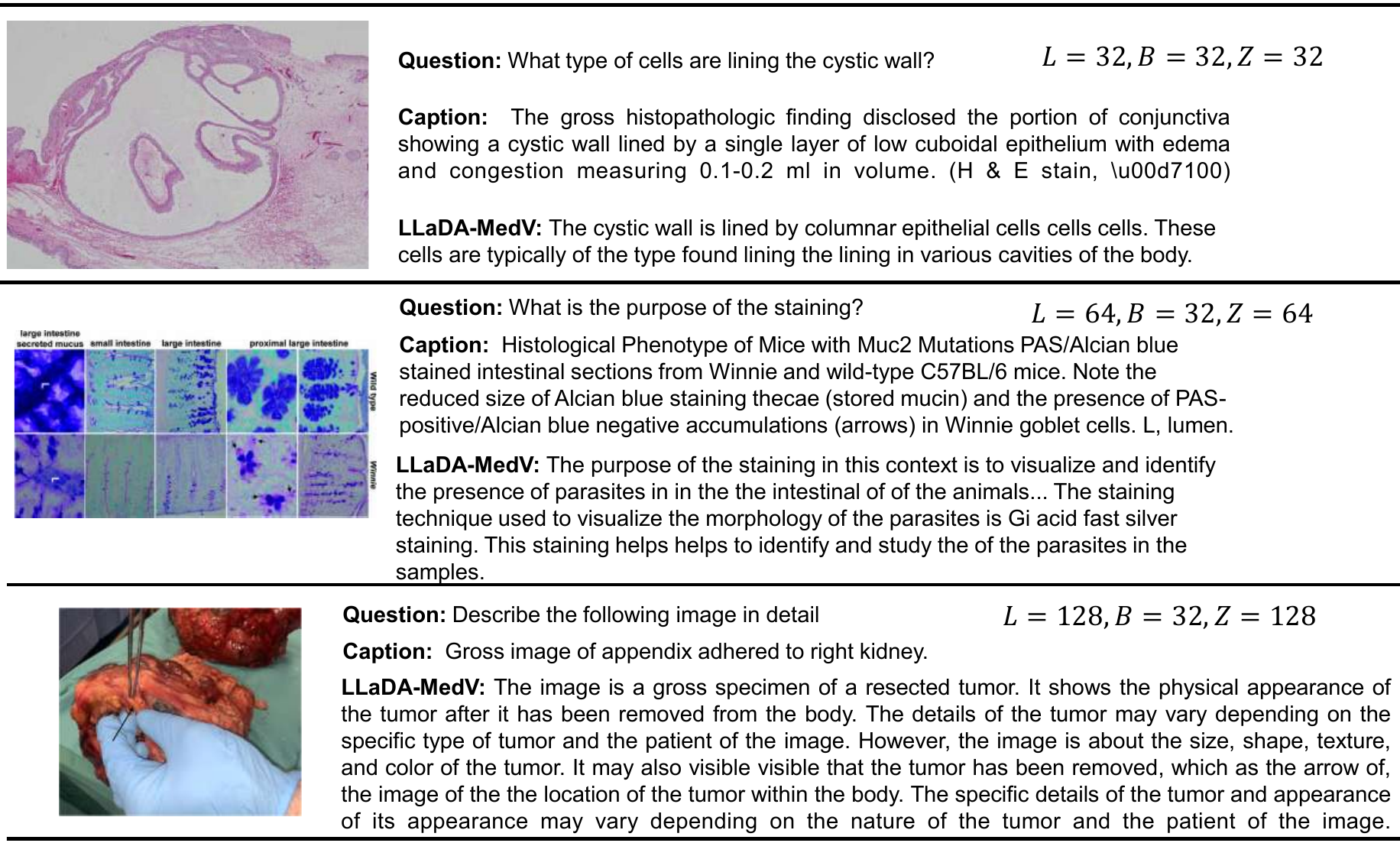}
  \caption{Illustration of LLaDA-MedV responses on open-end biomedical conversation as response length $L$ changes. Image, question and corresponding caption comes from ~\cite{li2023llava}. We omit GPT-4 reference answer for clarity. }
  \label{fig:inference-l}
\end{figure}
\begin{figure}[ht]
  \centering
  \includegraphics[width=1\linewidth]{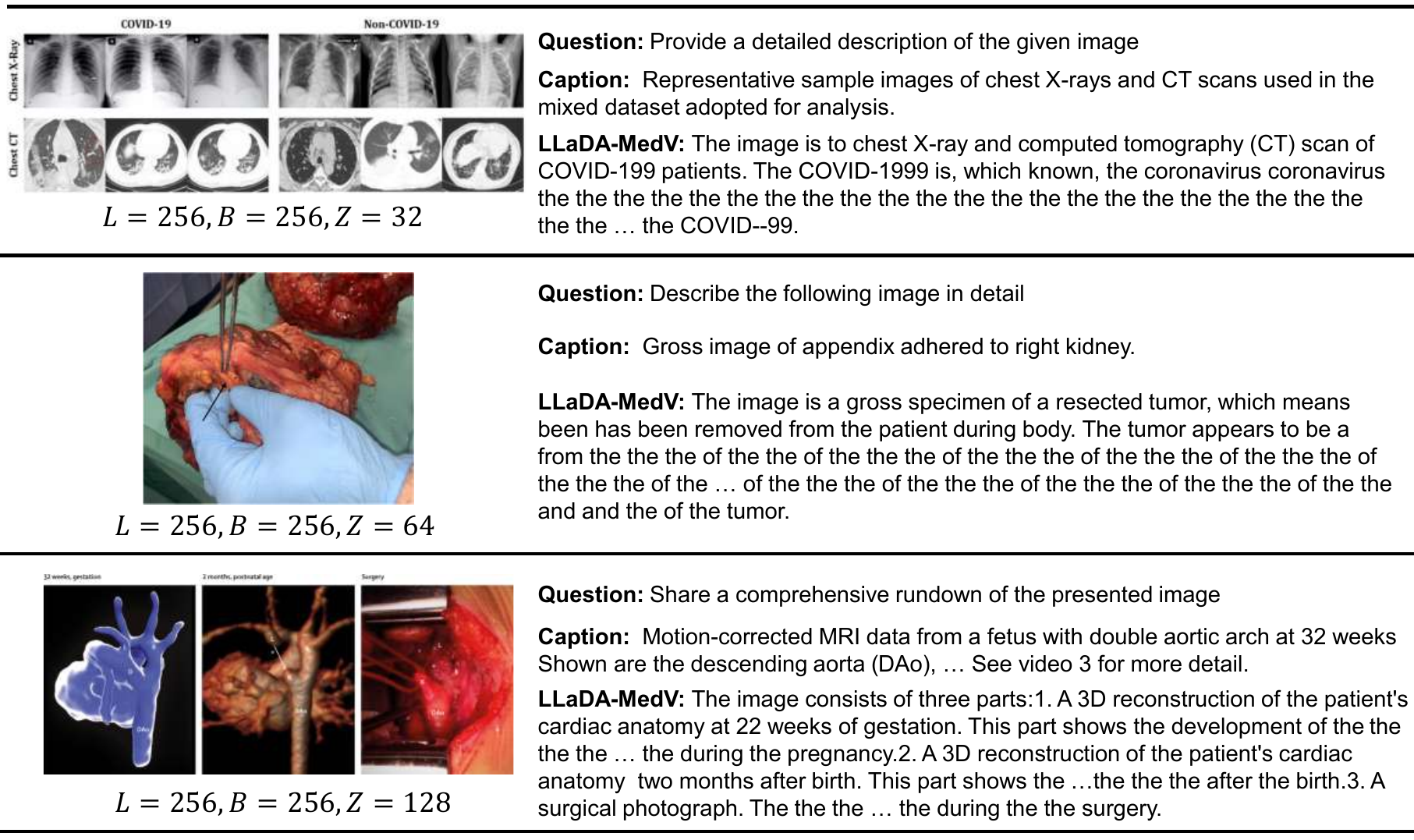}
  \caption{Illustration of LLaDA-MedV responses on open-end biomedical conversation as sampling steps $Z$ changes. Image, question and corresponding caption comes from ~\cite{li2023llava}. We omit GPT-4 reference answer for clarity. }
  \label{fig:inference-z}
\end{figure}
\begin{figure}[h]
  \centering
  \includegraphics[width=1\linewidth]{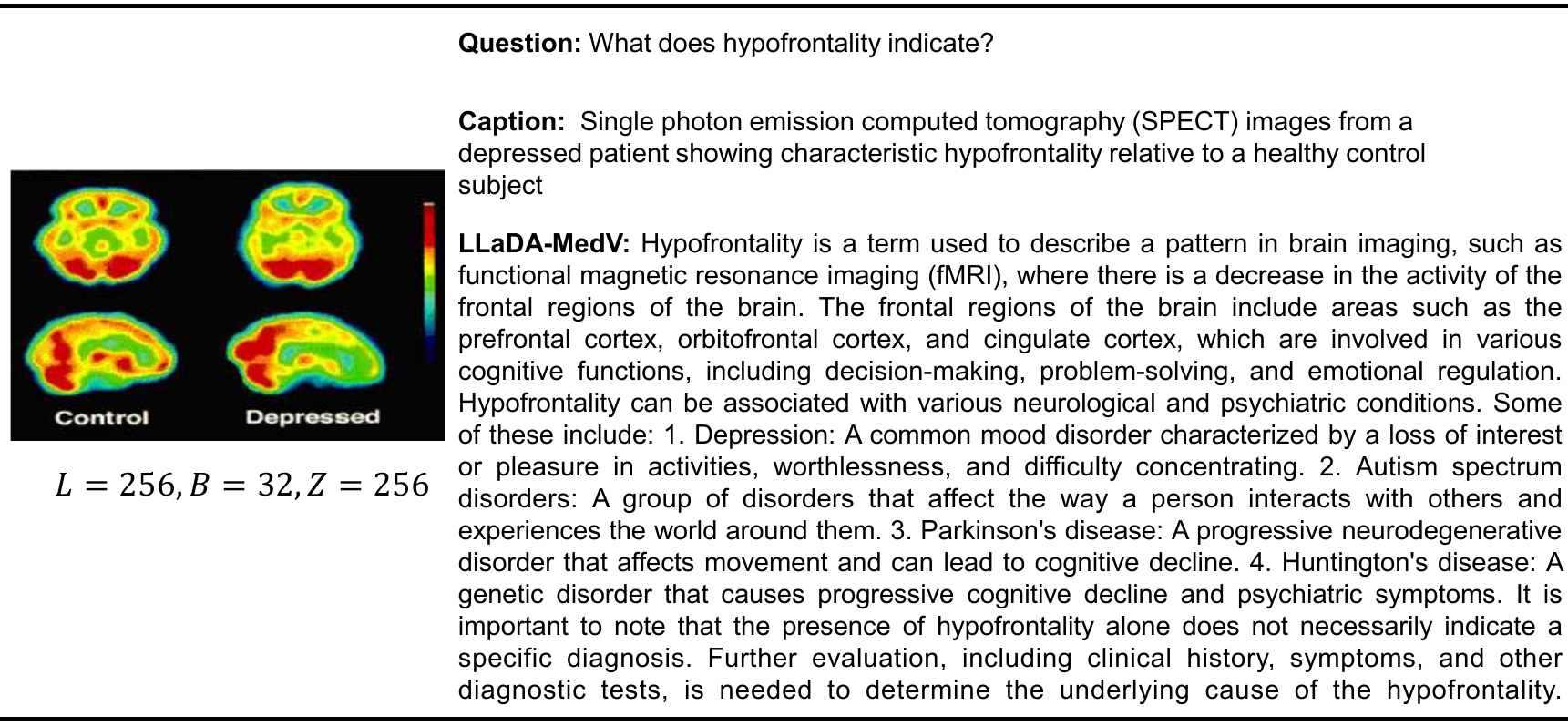}
  \caption{Illustration of LLaDA-MedV responses on open-end biomedical conversation as block length $B$ changes. Image, question and corresponding caption comes from ~\cite{li2023llava}. We omit GPT-4 reference answer for clarity. }
  \label{fig:inference-b1}
\end{figure}
\begin{figure}[h]
  \centering
  \includegraphics[width=1\linewidth]{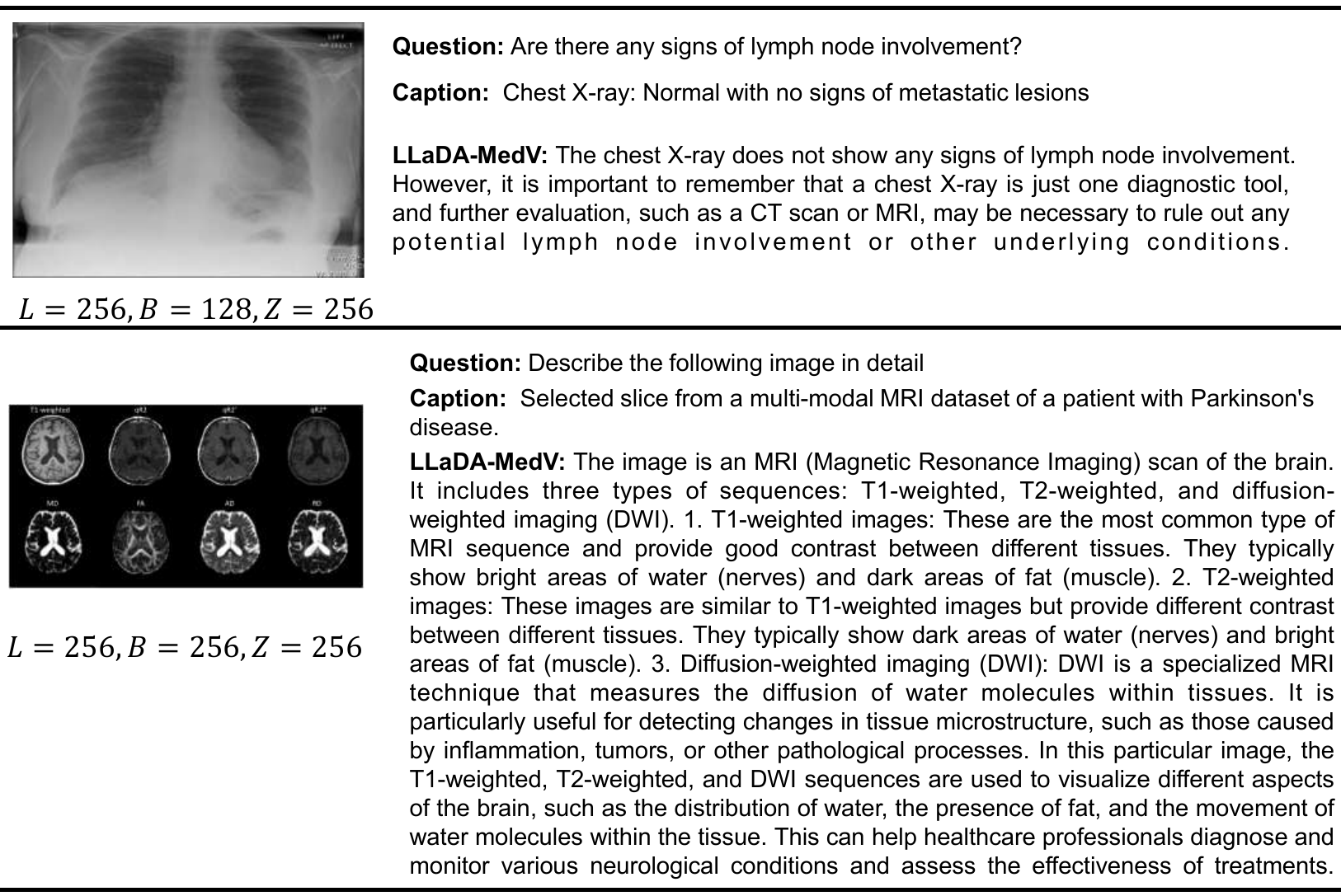}
  \caption{Illustration of LLaDA-MedV responses on open-end biomedical conversation as block length $B$ changes. Image, question and corresponding caption comes from ~\cite{li2023llava}. We omit GPT-4 reference answer for clarity. }
  \label{fig:inference-b2}
\end{figure}

Specifically, We observe that variations in the generation length $L$ impact overall performance in open-ended biomedical conversation tasks, as shown in Fig.~\ref{fig:inference-l}. This discrepancy may raise initial concerns about the fairness of our comparisons, since longer responses typically provide richer context to GPT-based evaluators, potentially leading to higher perceived answer quality. However, we emphasize that all evaluations are conducted under a controlled setting in which models are encouraged to generate responses of fixed length ($L = 256$). For ARMs such as LLaVA-Med, this constraint is enforced by setting the maximum token length to 256. In contrast, LLaDA-MedV leverages a masked prediction mechanism, which inherently encourages the generation of more complete and informative responses within the same length constraint. Additionally, differences in benchmark design and evaluation protocols may also contribute to the observed variation in performance trends. We further find that the number of sampling steps $Z$ plays an important role in controlling response diversity. As shown in Fig.~\ref{fig:inference-z}, when generating long responses, an insufficient number of steps can lead to noticeable token repetition. Lastly, when using semi-autoregressive generation, the choice of block length $B$ must be made with care. To aid understanding, we provide qualitative visualizations in Fig.~\ref{fig:inference-b1} and Fig.~\ref{fig:inference-b2}.

\end{document}